\documentclass[manuscript]{acmart}

\AtBeginDocument{%
  \providecommand\BibTeX{{%
    \normalfont B\kern-0.5em{\scshape i\kern-0.25em b}\kern-0.8em\TeX}}}

\setcopyright{acmcopyright}
\copyrightyear{2024}
\acmYear{2024}
\acmDOI{XXXXXXX.XXXXXXX}

\usepackage{xcolor}
\usepackage{hyperref}
\usepackage{cleveref}
\usepackage{multirow}
\usepackage{array}
\usepackage{graphicx}
\usepackage{pifont}

\newcommand\Tstrut{\rule{0pt}{2ex}}         
\newcommand\Bstrut{\rule[-1ex]{0pt}{0pt}}   
\newcolumntype{C}[1]{>{\centering}m{#1}}
\newcolumntype{E}[1]{>{\centering\arraybackslash}m{#1}}

\begin{document}

\hbadness=2000000000
\vbadness=2000000000
\hfuzz=100pt

\setlength{\headheight}{20.48303pt}

\title{Video Unsupervised Domain Adaptation with Deep Learning: A~Comprehensive~Survey}

\author{Yuecong Xu}
\email{yc.xu@nus.edu.sg}
\orcid{0000-0002-4292-7379}
\affiliation{%
  \institution{Department of Electrical and Computer Engineering, National University of Singapore}
  \streetaddress{4 Engineering Drive 3}
  \country{Singapore}
  \postcode{117583}
}

\author{Haozhi Cao}
\email{haozhi002@e.ntu.edu.sg}
\orcid{0000-0001-7703-3490}
\author{Lihua Xie}
\email{elhxie@ntu.edu.sg}
\orcid{0000-0002-7137-4136}
\affiliation{%
  \institution{School of Electrical and Electronic Engineering, Nanyang Technological University}
  \streetaddress{50 Nanyang Avenue}
  \country{Singapore}
  \postcode{639798}
}

\author{Xiaoli Li}
\email{xlli@i2r.a-star.edu.sg}
\orcid{0000-0002-0762-6562}
\author{Zhenghua Chen}
\email{chen0832@e.ntu.edu.sg}
\orcid{0000-0002-1719-0328}
\affiliation{%
  \institution{Institute for Infocomm Research, A*STAR}
  \streetaddress{1 Fusionopolis Way}
  \country{Singapore}
  \postcode{138632}
}

\author{Jianfei Yang}
\authornote{Corresponding author.}
\orcid{0000-0002-8075-0439}
\affiliation{%
  \institution{School of Mechanical and Aerospace Engineering and School of Electrical and Electronic Engineering, Nanyang Technological University}
  \streetaddress{50 Nanyang Avenue}
  \country{Singapore}
  \postcode{639798}
}
\email{jianfei.yang@ntu.edu.sg}

\renewcommand{\shortauthors}{Xu and Yang, et al.}

\begin{abstract}
    Video analysis tasks such as action recognition have received increasing research interest with growing applications in fields such as smart healthcare, thanks to the introduction of large-scale datasets and deep learning-based representations. However, video models trained on existing datasets suffer from significant performance degradation when deployed directly to real-world applications due to domain shifts between the training public video datasets (source video domains) and real-world videos (target video domains). Further, with the high cost of video annotation, it is more practical to use unlabeled videos for training. To tackle performance degradation and address concerns in high video annotation cost uniformly, the video unsupervised domain adaptation (VUDA) is introduced to adapt video models from the labeled source domain to the unlabeled target domain by alleviating video domain shift, improving the generalizability and portability of video models. This paper surveys recent progress in VUDA with deep learning. We begin with the motivation of VUDA, followed by its definition, and recent progress of methods for both closed-set VUDA and VUDA under different scenarios, and current benchmark datasets for VUDA research. Eventually, future directions are provided to promote further VUDA research. The repository of this survey is provided at \href{github.com/xuyu0010/awesome-video-domain-adaptation}{https://github.com/xuyu0010/awesome-video-domain-adaptation}.
\end{abstract}




\maketitle

\section{Introduction}
\label{section:intro}

With the rapid growth of video data at an extraordinary rate, automatic video analysis tasks, e.g., action recognition (AR) and video segmentation, have received increasing research interest with growing applications. Over the past decade, there have been great developments in various video analysis tasks. This is largely enabled by the emergence of diverse large-scale video datasets and the continuous advancement in video representation learning, particularly with deep neural networks and deep learning.

Despite the progress made in video analysis tasks (e.g., AR), most existing methods assume that the training and testing data are drawn from the same distribution, which yet may not hold in real-world applications. In practice, it is very common that the distribution of the training data from public datasets and testing data collected in real-world scenarios differ, and therefore a \textit{domain shift} between the training (source) and testing (target) domains exists. In these scenarios, we observe significantly degraded performances of trained video models in the testing (target) domain despite the strong capacity of deep neural networks. For instance, deep video models trained for autonomous driving with current datasets (e.g., KITTI, nuScenes) would not be applicable for nighttime autonomous driving; while deep video models trained with regular action recognition datasets (e.g., UCF101, HMDB51) may not be able to recognize actions of patients in hospitals. 

To tackle the performance degradation under domain shift, various \textit{domain adaptation} (DA) methods have been proposed to utilize labeled data in the source domain to execute tasks in the target domain. Domain adaptation methods generally aim to learn a model from the source domain that can be generalized to the target domain by minimizing the difference between domain distributions. Meanwhile, due to the high cost of annotating large-scale real-world data for deep learning, it is more feasible to obtain unlabeled data for models to be adapted to target domains. The \textit{unsupervised domain adaptation} (UDA) task is therefore introduced where models are adapted from the labeled source domain towards the unlabeled target domain by alleviating the negative effect of domain shift while avoiding costly data annotation.

While UDA with deep learning greatly improves the generalizability and portability of models by tackling domain shift, prior research for visual applications generally focused on image data. An intuitive method for \textit{video unsupervised domain adaptation} (VUDA) is to extend UDA methods for images to videos by directly substituting the image feature extractor with a video feature extractor (e.g., substituting 2D CNNs with 3D CNNs). Meanwhile, video representations obtained through conventional video feature extractors are mainly from spatial features. However, videos contain not only spatial features but also temporal features as well as features of other modalities, e.g., optical flow and audio features. Domain shift would occur for all of these features. Therefore, such a vanilla substitution strategy that ignores domain shift across the multiple modalities of features produces inferior results for VUDA.

Subsequently, various VUDA methods have been proposed to explicitly deal with the issue of domain shift for video tasks.
Based on how source and target domains are aligned, the VUDA methods can be generally categorized into five categories: a) adversarial-based methods~\cite{jamal2018deep,chen2019temporal,chen2022multi}, where feature generators are trained jointly with additional domain discriminators in an adversarial manner, with domain-invariant features obtained if the domain discriminators failed to discriminate whether they originate from the source or target domains; b) discrepancy-based (or metric-based) methods~\cite{jamal2018deep,gao2020pairwise}, where the discrepancy between the source and target domains are explicitly computed, while the target domain is aligned with the source domain by applying metric learning approaches, optimized with metric-based objectives such as MDD~\cite{zhang2019bridging}, CORAL~\cite{sun2016return}, and MMD~\cite{long2015learning};
c) semantic-based methods~\cite{song2021spatio,kim2021learning,sahoo2021contrast}, where domain-invariant features are obtained subject to certain semantic constraints by leveraging approaches such as mutual information maximization~\cite{torkkola2003feature}, clustering~\cite{xu2005survey}, contrastive learning~\cite{khosla2020supervised,chen2020simple}, and pseudo-labeling~\cite{lee2013pseudo,arazo2020pseudo}; d) reconstruction-based methods~\cite{wei2022unsupervised}, where domain-invariant features are extracted by encoders trained with data-reconstruction objectives with the network commonly structured as an encoder-decoder;
and e) composite methods~\cite{gao2020pastn,choi2020shuffle,xu2021aligning}, where domain-invariant features are extracted by optimizing a combination of different objectives (i.e., domain discrepancy objectives, adversarial objectives, and semantic-based objectives). It is possible to further divide the relatively ambiguous semantic-based and composite-based methods into sub-categories based on the specific semantic information leveraged or how different alignment objectives are combined.
However, thanks to the limited available research for VUDA, further sub-categorizing would result in trivial categories containing only few or even single available approach. As such, we stick to a broader categorization strategy to enable a broader picture of the current progress of VUDA approaches.

\begin{figure*}[t]
  \centering
  \includegraphics[width=1.\linewidth]{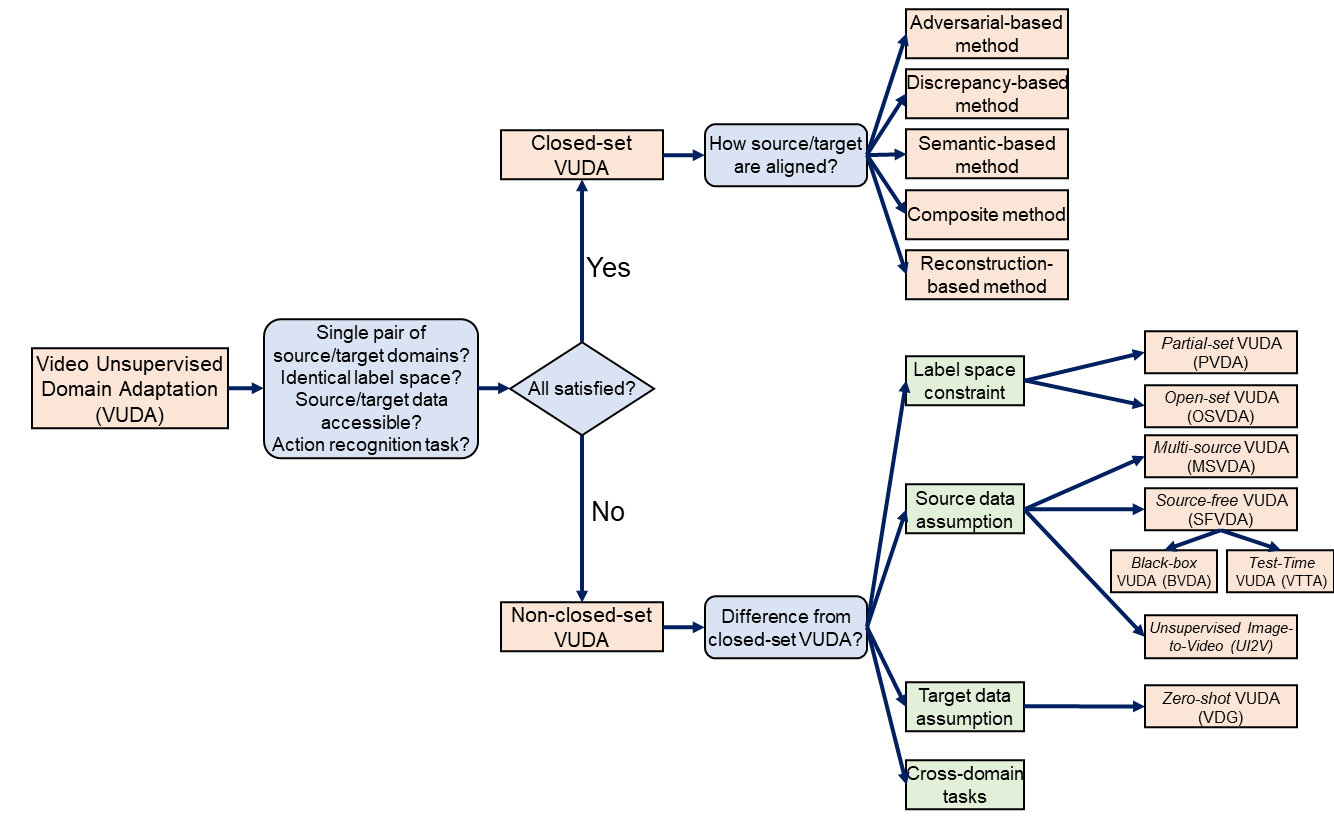}
  \caption{
  Overview of the categorization of the different VUDA methods. Closed-set VUDA methods are constrained by the constraint of an identical label space shared by the single pair of video source/target domains and assume that both the source and target data are accessible, with action recognition as the cross-domain task. Any VUDA methods that does not satisfy the four constraints/assumptions are considered as non-closed-set VUDA. Closed-set VUDA methods can be categorized into five categories based on how source and target domains are aligned. A broader categorization strategy is adopted thanks to the limited available research, and to enable a broader picture of the current progress of VUDA approaches. Non-closed-set VUDA methods are categorized into four categories by how their related scenarios differ from the closed-set VUDA.
  }
  \label{figure:1-1-categorization}
\end{figure*}

Meanwhile, though the aforementioned VUDA methods enable the learning of transferable knowledge across video domains, they are built upon several constraints and assumptions. These include the constraint of an identical label space shared by the single pair of the video source and target domains and the assumption that both the labeled source and unlabeled target domain data are accessible during training, with action recognition as the cross-domain task. VUDA under these constraints and assumptions are denoted as \textit{closed-set VUDA}, and may not hold in real-world scenarios, prompting concerns about model portability due to issues such as data security.
Over the past few years, there have been various scattered research that looks to differ the constraints and assumptions such that VUDA methods could be more applicable in real-world scenarios. These methods could be broadly categorized based on the difference over the targeted setting against the closed-set setting as: a) methods with different label space constraint; b) methods with different source data assumption; c) methods with different target data assumption; and d) methods with different cross-domain tasks. Fig.~\ref{figure:1-1-categorization} presents an overview of the categorization of the different VUDA methods.

There had been prior surveys focusing on shallow and deep DA and UDA approaches with their applications in various image and natural language processing tasks. For example, \cite{beijbom2012domain,patel2015visual} surveyed shallow DA approaches for image tasks while~\cite{csurka2017domain} also briefly recapped some deep DA approaches. Subsequently, \cite{wang2018deep} further summarized other deep DA approaches for image tasks. Later, \cite{kouw2019review} outlined various UDA methods while \cite{wilson2020survey} focused on UDA methods with deep learning. Several other works~\cite{bungum2011survey,chu2020survey,ramponi2020neural} discussed DA and UDA for various natural language processing tasks, such as machine translation and sentiment analysis. Meanwhile, there were also works surveyed on the broader transfer learning (TL) topic~\cite{pan2009survey,weiss2016survey,tan2018survey,zhuang2020comprehensive}, where domain adaptation can be viewed as a special case. Despite the effort made in surveying comprehensively DA, UDA, and TL, there has been no specific survey that investigates UDA for video tasks (i.e., VUDA). To the best of our knowledge, this is the \textit{first} article that investigates and summarizes the recent progress of video unsupervised domain adaptation, where current works are generally deep learning-based. By summarizing existing literature, we propose the prospect of VUDA and the direction of future VUDA research.

The rest of the paper is organized as follows. Section~\ref{section:def} defines the closed-set VUDA specifically and introduces the relevant notations. In Section~\ref{section:closed-methods}, we review the deep learning-based closed-set VUDA methods, while methods for VUDA under different constraints, assumptions, and tasks are reviewed in Section~\ref{section:other-methods}. We further summarize existing cross-domain video datasets for benchmarking VUDA in Section~\ref{section:datasets}. Insights and future directions of VUDA research are discussed in Section~\ref{section:discussion} and the paper is concluded in Section~\ref{section:concl}.

\section{Definitions and Notations}
\label{section:def}

\begin{figure*}[t]
  \centering
  \includegraphics[width=1.\linewidth]{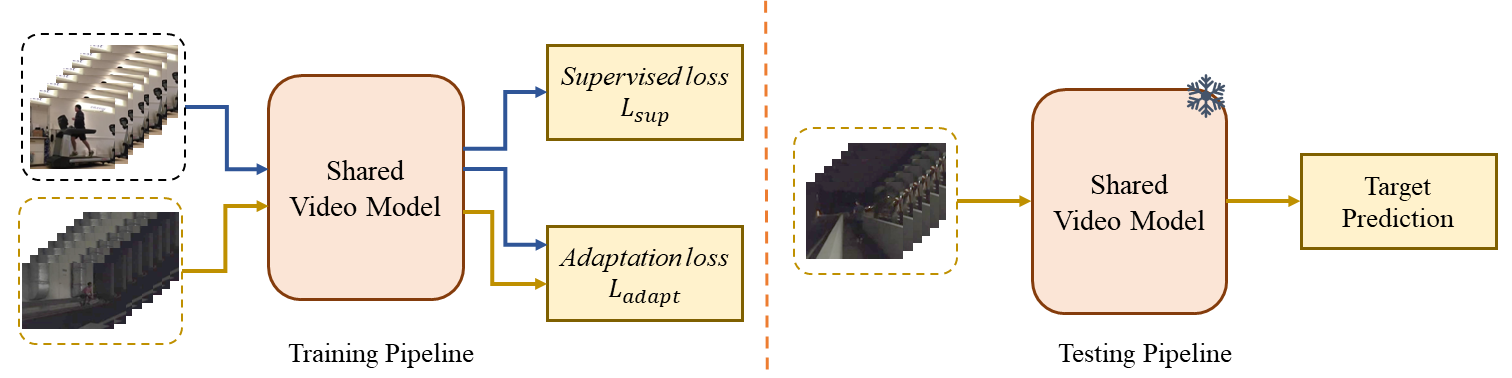}
  \caption{Training and testing pipeline of VUDA. Note the {\color{blue}blue} lines indicate the data flow for the labeled source data, while the {\color{olive}olive} lines indicate the data flow for the unlabeled target data. The trained shared video model is frozen during testing (indicated by \ding{100}).}
  \label{figure:2-a-pipeline}
\end{figure*}

In this section, we define the video unsupervised domain adaptation (VUDA) and introduce the relevant notations used in this survey. To maintain consistency with current VUDA works, the definitions and notations are defined with reference to \cite{xu2021aligning,xu2021partial}. In VUDA, we are given a collection of $M_{S}$ video source domains and $\{\mathcal{D}_{S}^{1}, \mathcal{D}_{S}^{2}, ... , \mathcal{D}_{S}^{M_{S}}\}$ (which may or may not be accessible) and a collection of $M_{T}$ video target domains $\{\mathcal{D}_{T}^{1}, \mathcal{D}_{T}^{2}, ... , \mathcal{D}_{T}^{M_{T}}\}$.
Each source domain contains $N_{S}^{k}$ labeled videos $\mathcal{D}_{S}^{k}=\{(V_{S}^{k,i},y_{S}^{k,i})\}^{N_{S}^{k}}_{i=1}$, characterized by the underlying probability distribution $p^{k}$ associated with the label space $\mathcal{Y}_{S}^{k}$ that contains $|\mathcal{C}_{S}^{k}|$ classes. Meanwhile each target domain contains $N_{T}^{r}$ unlabeled videos $\mathcal{D}_{T}^{r}=\{V_{T}^{r,i}\}^{N_{T}^{r}}_{i=1}$ characterized by the underlying probability distribution $q^{r}$ associated with the label space $\mathcal{Y}_{T}^{r}$ that contains $|\mathcal{C}_{T}^{r}|$ classes.
The goal of VUDA is to design a target model $G_{T}(.;\theta_{T})$ for the target domain that originates or is initialized from the source model $G_{S}(.;\theta_{S})$, which is capable of learning transferable features from the labeled source domains and minimize the empirical target risk $\epsilon_{T}$ across all target domains performed on certain video-based tasks (e.g., human action recognition, video segmentation or video quality assessment). The same task is performed on both the video source and target domains. An overall training and testing pipeline of VUDA is presented in Fig.~\ref{figure:2-a-pipeline}.

The domain adaptation theory~\cite{ben2010theory} proves that the empirical target risk $\epsilon_{T}$ is upper-bounded by three terms: a) the combined error of the ideal joint hypothesis on the source and target domains, which is assumed to be small so that domain adaptation can be achieved; b) the empirical source-domain error, and c) the divergence measured between source and target domains. All VUDA methods attempt to minimize $\epsilon_{T}$ by minimizing the third term and/or the second term which would be discussed in detail in subsequent Sections.

While the above definition can be viewed as a general scenario of VUDA, it could be too challenging to tackle. Therefore, existing works would normally apply certain constraints or assumptions towards the general scenario to form scenarios that could be better tackled. The most common scenario is the closed-set VUDA scenario, which sets the below constraints and assumptions:
\begin{itemize}
    \item There would be only $M_{S}=1$ video source domain and only $M_{T}=1$ target domain (superscripts $k$ and $r$ are therefore omitted for simplicity).
    \item Both the source domain videos $\mathcal{D}_{S}$ and the source model $G_{S}(.;\theta_{S})$ are accessible.
    \item The source and target domain videos share the same label space (i.e.,~$\mathcal{Y}_{S}=\mathcal{Y}_{T}$ and $|\mathcal{C}_{S}|=|\mathcal{C}_{T}|$).
    \item The domain shift between the source and target domains is categorized as covariate shift~\cite{sugiyama2007covariate}, where $p(V_{S})\neq p(V_{T})$ but $p(y_{S}|V_{S})=p(y_{T}|V_{T})$.
    \item The source and target domain videos share the same model (i.e.,~$G_{S}(.;\theta_{S})=G_{T}(.;\theta_{T})$.
    \item The video-based task to be performed is assumed to be the human action recognition task.
\end{itemize}
Based on the constraints and assumptions for the closed-set VUDA scenario, notations would be simplified while the superscripts $k, r$ are omitted and the joint label space $\mathcal{Y}$ contains $\mathcal{C}$ classes. 
\section{Methods for Closed-Set Video Unsupervised Domain Adaptation}
\label{section:closed-methods}

In this section, we review the various deep learning-based \textit{closed-set} VUDA methods whose training and testing follow the constraints and assumptions as presented in Section~\ref{section:def}. As briefly mentioned in Section~\ref{section:intro}, deep learning-based closed-set VUDA methods could be generally categorized into five categories. We discuss each category in the following sections. 
The list of all reviewed closed-set VUDA methods is collated and compared in Tab.~\ref{table:3-closed_set-methods}.

\begin{table*}[ht]
\center
\caption{Different categories of methods for closed-set VUDA. Methods are listed in chronological order.}
\begin{tabular}{m{.15\linewidth}|C{.4\linewidth}|E{.35\linewidth}}
    \hline
    \hline
    Categories & Brief Descriptions & Example Methods \Tstrut\Bstrut\\
    \hline
    Adversarial-based & Domain discriminators to encourage domain confusion through adversarial objectives across video domain.\Tstrut\Bstrut & DAAA~\cite{jamal2018deep}, TA\textsuperscript{3}N~\cite{chen2019temporal},  TCoN~\cite{pan2020adversarial}, MM-SADA~\cite{munro2020multi}, MA\textsuperscript{2}L-TD~\cite{chen2022multi}, CIA~\cite{yang2022interact}, ABG~\cite{luo2020adversarial} \Tstrut\Bstrut\\
    \hline
    Discrepancy-based & Discrepancy between domains are explicitly computed, align domains by applying metric learning approaches.\Tstrut\Bstrut & AMLS~\cite{jamal2018deep}, PTC~\cite{gao2020pairwise} \Tstrut\Bstrut\\
    \hline
    Semantic-based & Domain-invariant features are obtained by exploiting the shared semantics across domains.\Tstrut & STCDA~\cite{song2021spatio}, CMCo~\cite{kim2021learning}, CoMix~\cite{sahoo2021contrast}, CO\textsuperscript{2}A~\cite{da2022dual}, A\textsuperscript{3}R~\cite{zhang2022audio}, DVM~\cite{wu2022dynamic} \Tstrut\Bstrut\\
    \hline
    Composite & Exploit a composite of approaches to capitalize on the strength of each approach.\Tstrut
    & NEC-Drone~\cite{choi2020unsupervised}, SAVA~\cite{choi2020shuffle}, PASTN~\cite{gao2020pastn}, MAN~\cite{gao2021novel}, ACAN~\cite{xu2021aligning}, CVTD~\cite{choi2022self}, GLAD~\cite{lee2024glad} \Tstrut\Bstrut\\
    \hline
    Reconstruction-based & Domain-invariant features from encoder-decoder networks with data-reconstruction objectives.\Tstrut & TranSVAE~\cite{wei2022unsupervised} \Tstrut\Bstrut\\
    \hline
    \hline
\end{tabular}
\label{table:3-closed_set-methods}
\end{table*}

\subsection{Adversarial-based VUDA Methods}
\label{section:closed-methods:adversarial}
With the assumption that the source empirical risk would be small with supervised learning on source videos, methods have been proposed to achieve effective domain adaptation by minimizing the discrepancy between the source and target domains. Intuitively, if the source and target domains share the same data distribution, the domain discriminators would be unable to identify whether a video originates from the source or target domain. Such intuition, along with the success of Generative Adversarial Networks (GANs)~\cite{goodfellow2014generative}, motivates the proposal of adversarial-based VUDA methods, where additional domain discriminators are leveraged to encourage domain confusion through adversarial objectives across the source and target video domains. The discrepancy between the source and target domains is therefore minimized implicitly. Adversarial-based domain adaptation methods have previously found their success in image-based UDA tasks (e.g.,~image recognition~\cite{ganin2015unsupervised,kang2020effective,yang2020mind,yang2021robust}, object detection~\cite{chen2018domain,cai2019exploring} and person re-identification~\cite{yang2020part}), therefore it is intuitive to extend adversarial-based methods to VUDA.

One primitive work in this category is the Deep Adversarial Action Adaptation (DAAA)~\cite{jamal2018deep}, where the original imaged-based DANN~\cite{ganin2015unsupervised,ganin2016domain} is extended to videos by substituting the image extractor (2D-CNN~\cite{gu2018recent}) with the clip/video extractor (3D-CNN~\cite{ji20123daction}) while the input is changed from images to clips, which are formed by sampling frames from the videos. Though achieving much better performances compared to shallow domain adaptation methods, DAAA ignores the difference between the source-target discrepancies of spatial and temporal features, adapting spatial and temporal features uniformly and indiscriminately. Subsequently, Temporal Attentive Adversarial Adaptation Network (TA\textsuperscript{3}N)~\cite{chen2019temporal} leverages on Temporal Relation Network (TRN)~\cite{zhou2018temporal} to obtain more explicit temporal features, and align videos with both spatial and temporal features. TA\textsuperscript{3}N is designed to focus dynamically on aligning the temporal dynamics which have higher domain discrepancy to effectively align the temporal features of videos. To achieve this, TA\textsuperscript{3}N adopts an adaptive non-parametric attention mechanism based on domain prediction entropy. The strategy of aligning and adapting spatial and temporal features separately is further extended to the alignment of multiple levels of temporal features separately in MA\textsuperscript{2}L-TD~\cite{chen2022multi}, with each level of temporal feature corresponding to the different length of video segment/clip generated by a temporally-dilated feature aggregation module. MA\textsuperscript{2}L-TD assigns attention weights to the alignment of different levels by the degree of domain confusion in each level, where larger weights are assigned to levels where the corresponding domain discriminator cannot classify domains correctly. 

Meanwhile, videos contain a series of non-key frames whose noisy background information is unrelated to the action and could affect adaptation negatively. Temporal Co-attention Network (TCoN)~\cite{pan2020adversarial} copes with non-key frames by selecting the key segments that are critical for cross-domain action recognition. TCoN selects key segments by computing attention scores of each segment based on action informativeness and cross-domain similarity, obtained by a self-attention-inspired cross-domain co-attention matrix.
Instead of direct adaptation across the source and target video features, TCoN adapts target video features to the source ones by constructing target-aligned source video features via transforming the original source video features through the cross-domain co-attention matrix. Lately, graphs have also been leveraged for aligning video domains, where the ABG~\cite{luo2020adversarial} is proposed to model the temporal correlations across the domains with a network topology of a bipartite graph. ABG then adopts the adversarial alignment strategy similar to prior methods.

Besides spatial and temporal features which are generally obtained from the RGB modality, videos also contain information about other modalities, such as optical flow and audio modalities. The multi-modality nature of videos poses more challenges towards VUDA as domain shift would be incurred for each modality. Methods have therefore been proposed to align source and target videos leveraging on the multi-modality information. Among these, MM-SADA~\cite{munro2020multi} leverages the RGB and optical flow modalities, where adversarial alignment is applied to each modality separately. MM-SADA further adopts self-supervision learning across different modalities to learn the temporal correspondence between the different modalities. More recently, Cross-modal Interactive Alignment (CIA)~\cite{yang2022interact} aligns video features with RGB, optical flow, and audio modalities. CIA further observes that cross-modal alignment could conflict with cross-domain alignment in VUDA, therefore it enhances the transferability of each modality by cross-modal interaction through a Mutual Complementarity (MC) module. The different modalities are therefore refined by absorbing the transferable knowledge from other modalities before they are aligned across source and target domains.

\subsection{Discrepancy-based VUDA Methods}
\label{section:closed-methods:discrepancy}
While adversarial-based VUDA methods achieve decent performances in various VUDA benchmarks, the above methods do not compute the discrepancy between source and target domains or measure such discrepancy implicitly. Moreover, previous studies~\cite{saxena2021generative,gonog2019review,wang2017generative} have shown that adversarial training is unstable and may lead to model collapse and non-convergence. Discrepancy-based VUDA methods are therefore proposed as the more intuitive and stable approach towards tackling VUDA by computing and minimizing the video domain discrepancy explicitly. An early method is as proposed in AMLS~\cite{jamal2018deep} where the target videos are modeled as a sequence of points on the Grassmann manifold~\cite{turaga2008statistical} with each point corresponding to a collection of clips aligned temporally, and the source videos are modeled as a single point on the manifold. VUDA is tackled by minimizing the Frobenius norm~\cite{huckle2007frobenius} between the source point and the series of target points on the Grassmann manifold. Meanwhile, later method has also leveraged multi-modal information with discrepancy computation and minimization. The Pairwise Two-stream ConvNets (PTC)~\cite{gao2020pairwise} minimizes the MMD~\cite{long2015learning} loss across both RGB and optical flow modalities achieving better performances on more complex domain shift scenarios. PTC further improves its generalizability by fusing the RGB and optical flow features through a self-attention weight mechanism and selection of training videos at the boundary of action classes through a sphere boundary sample-selecting scheme.

\subsection{Semantic-based VUDA methods}
\label{section:closed-methods:semantic}
Besides minimizing discrepancies explicitly by discrepancy-based methods and implicitly by adversarial-based methods, aligning source and target video domains can also be accomplished by semantic-based VUDA methods~\cite{yang2021advancing} which exploit the shared semantics across the source and target domains such that domain-invariant features are obtained. Intuitively, if the target video domain aligns well with the source video domain through a certain model, the model extracts source-like representations for target videos, semantics embedded within the source video domain should therefore be shared with the target domain. 
Typical implications of shared semantics across domains include: a) spatio-temporal association: frames and clips (under different modalities) of videos possess strong spatio-temporal association and are placed in the correct time order and pose; b) feature clustering: features related to the videos of the same action classes are clustered and close to each other, whereby features related to videos of different action classes are placed further away; and c) modality correspondence: features extracted from the different modalities of the same video are close together.

\textbf{a) Methods Leveraging Spatio-Temporal Association.}
One typical semantic-based method proposed is the Spatio-Temporal Contrastive Domain Adaptation (STCDA)~\cite{song2021spatio}, which mines video representations of both RGB and optical flow modalities by applying a contrastive loss on both the clip and video level such that frames and clips are spatially and temporally associated. STCDA further bridges the domain shift of source and target videos by a video-based contrastive alignment (VCA) which minimizes the distance of the intra-class source and target features and maximizes the distance of the inter-class source and target features on the Reproducing Kernel Hilbert Space (RKHS)~\cite{berlinet2011reproducing}. The labels of target videos are obtained by pseudo-labeling through clustering with the features of the labeled source videos. Contrastive learning has also been applied in CMCo~\cite{kim2021learning} which aims to extract video features with modality correspondence across RGB and optical flow modalities. Similarly, CoMix~\cite{sahoo2021contrast} enforces temporal speed invariance in videos which encourages features extracted from the same video yet sampled with different temporal speeds to be similar. CoMix further employs the supervised contrastive loss~\cite{khosla2020supervised} to target data by computing pseudo-labels and selecting target data whose pseudo-labels are of high confidence. Apart from applying contrastive learning, another possible approach that leverages spatio-temporal association for semantic-based VUDA concerns the use of self-supervised learning~\cite{zhai2019s4l}, which relies on surrogate tasks that can be formulated using only unsupervised data. The surrogate tasks are designed to help models learn visual representation, one typical task is to predict image rotation angle~\cite{gidaris2018unsupervised}. Prior works~\cite{zhai2019s4l} have also leveraged on exemplar self-supervision technique~\cite{dosovitskiy2014discriminative} which aims to learn visual representation that is invariant to image transformations such as cropping, HSV color randomization, and horizontal mirroring.

\textbf{b) Methods Leveraging Feature Clustering.}
Contrastive learning can also be leveraged with the goal of feature clustering, with CO\textsuperscript{2}A~\cite{da2022dual} being a typical example where it employed contrastive learning on both the clip and video levels. CO\textsuperscript{2}A also introduced supervised contrastive learning~\cite{khosla2020supervised} for source video feature learning. Furthermore, CO\textsuperscript{2}A encourages coherent correspondence predictions between source/target video pairs. The correspondence predictions of the source/target video pair predict whether the source/target videos are of the same label, and are obtained from either the label/pseudo-label or the features of source/target videos trained with contrastive learning. Besides contrastive learning, mixing source and target domain samples (or equivalently leveraging MixUp~\cite{zhang2018mixup} across the source and target domains) have proven to benefit unsupervised domain adaptation for image-based tasks~\cite{xu2020adversarial,yan2020improve,wu2020dual} and improve model robustness. To further exploit shared action semantics, CoMix~\cite{sahoo2021contrast} adopts such a strategy that incorporates synthetic videos into its contrastive objective. The synthetic videos are obtained by mixing the background of the video from one domain with the video from another domain. More recently, DVM~\cite{wu2022dynamic} leverages MixUp to address the domain-wise gap directly at the input level, where the target videos are fused with the source videos progressively on the pixel-level. The corresponding target videos of the source videos are selected by obtaining the pseudo-labels of the target videos and matching them with the given labels of the source videos.

\textbf{c) Methods Leveraging Modality Correspondence.}
The above methods mostly deal with the domain gap between source and target videos with RGB and/or optical flow modalities. The A\textsuperscript{3}R~\cite{zhang2022audio} observe that sounds of actions can act as natural domain-invariant cues. Unlike previous methods where pseudo-labels are obtained directly by applying a classifier to the RGB and/or optical flow input, A\textsuperscript{3}R introduces an absent activity learning where audio predictions are leveraged to indicate which actions cannot be heard in the video, while visual predictions are further encouraged to have low probabilities for these ‘pseudo-absent’ actions. A\textsuperscript{3}R further proposes audio-balanced learning which exploits audio in the source domain to cluster samples. Finally, A\textsuperscript{3}R applies an audio-balanced loss where the rare actions are weighted higher to handle the semantic shift between domains.

\subsection{Reconstruction-based VUDA Methods}
\label{section:closed-methods:reconstruct}
Reconstruction-based VUDA methods deal with VUDA by obtaining domain-invariant features by an encoder-decoder network trained with data-reconstruction objectives. There had been some image-based domain adaptation works leveraging the reconstruction-based approach~\cite{ghifary2016deep,yang2020label,deng2021deep} thanks to its robustness to noise. However, there are few attempts in extending the reconstruction-based approach to VUDA due to the complexity of video reconstruction. TranSVAE~\cite{wei2022unsupervised} is a recent attempt in leveraging data-reconstruction objectives for VUDA. It aims to disentangle domain information from other information during adaptation by disentangling the cross-domain videos into domain-specific static variables and domain-invariant dynamic variables. With domain information obtained, the effect of domain discrepancy on the prediction task could be largely eliminated. The disentanglement is achieved through a Variational AutoEncoder (VAE)-structured~\cite{kingma2019introduction} network which models the cross-domain video generation process. TranSVAE further ensures the disentanglement serves the adaptation purpose by applying objectives to constraint the latent factors during the disentanglement, such as minimizing mutual dependence across static and dynamic variables and applying task-specific supervision on the dynamic variable from the source domain.

\subsection{Composite VUDA Methods}
\label{section:closed-methods:composite}
The above categories attempt to tackle VUDA from different perspectives, and all have their own strength and shortcomings. The adversarial-based approach is the more common approach thanks to its high performance and ease of implementation, yet it relies on unstable adversarial learning which may result in fragile convergence and requires additional domain discriminators during training. The divergence-based approach computes domain discrepancies explicitly without additional network components, and its optimization is more stable. However, it generally produces inferior performances compared to either the adversarial-based approach or the semantic-based approach. The semantic-based approach also results in high adaptation performances and can be extended to other scenarios of adaptation (e.g., source-free VUDA), but they are susceptible to noise and are optimized with higher computation cost. 

To capitalize on the strength of each approach for a more effective VUDA, various VUDA methods exploit a composite of approaches. For example, the NEC-Drone~\cite{choi2020unsupervised} proposed to combine the adversarial-based approach with the semantic-based approach by applying a triplet loss on the source data to learn embeddings of the videos which are agnostic of the specific classes but are aware of being similar. Similarly, the Pairwise Attentive adversarial Spatio-Temporal Network (PASTN)~\cite{gao2020pastn} also exploits both the adversarial-based and semantic-based approaches. PASTN is designed as a pairwise network with dual domain discriminators, one of which is structured without backpropagation and outputs the transferability weights for attentive adversarial learning. A margin-based discrimination loss~\cite{wu2017sampling} is employed across the source and target video features instead of the contrastive loss to compress intra-class samples within a margin and push inter-class samples away. This could extract shared semantics across source and target video domains by promoting feature clustering while taking the intra-class data distribution into consideration. 
Other methods that adopt the same combination of approaches are SAVA~\cite{choi2020shuffle}, CVTD~\cite{choi2022self}, and GLAD~\cite{lee2024glad}. SAVA aligns source and target video domains adversarially while attending to more discriminative clips through an attention mechanism. Further, SAVA focuses on encouraging temporal association in videos by applying an auxiliary clip order prediction task, which is more efficient and computationally less intensive than applying a contrastive loss. Meanwhile, CVTD applies a simple prediction task that predicts the difference of sampled temporal distances between the source and target videos. A standard adversarial domain classifier is also utilized by CVTD for further feature alignment. The more recent GLAD addresses background bias by performing the auxiliary temporal order prediction task. GLAD also proposes to tackle the temporal shift by utilizing temporal view alignment methods across both global and local views of videos. Each temporal view alignment method employs a dedicated adversarial domain classifiers to align the source and target features.

Besides combining the adversarial-based approach with the semantic-based approach, there have been other works that combine the adversarial-based approach with the discrepancy-based approach. For example, the Multiple-view Adversarial learning Network (MAN)~\cite{gao2021novel} performs adversarial learning to obtain domain-invariant video features from both RGB and optical flow modalities, fused by a Self-Attention Fusion Network (SAFN). MAN further improves domain invariance by applying the MK-MMD~\cite{long2015learning} loss over the fused video features. The Adversarial Correlation Alignment Network (ACAN)~\cite{xu2021aligning} is the other VUDA method that tackles VUDA with the composition of the adversarial-based and the discrepancy-based approach. Besides aligning spatial and temporal features, ACAN proposes to align correlation features extracted as long-range dependencies of pixels across spatiotemporal dimensions~\cite{wang2018non} by applying the adversarial domain loss to both the spatiotemporal video features and the correlation features. ACAN further aligns the correlation features by aligning the joint distribution of correlation information, which is computed as the covariance of correlation information. This is achieved by minimizing the Pixel Correlation Discrepancy across the source and target video domains implemented as the distance of correlation information distribution on the RKHS.

\subsection{Other Possible VUDA Methods}
\label{section:closed-methods:others}
Besides the aforementioned categories of methods, we observe that there are various approaches which have been explored in image-based UDA tasks and have proven their efficacy, but have not been explored for closed-set VUDA. These include methods that tackles domain shift from a causal point of view~\cite{zhang2015multi,yue2021transporting}, abstracting the UDA problem into causal models and facilitates UDA from the extracted causal information. Instead of obtaining domain-invariant features, causal methods~\cite{zhang2015multi,yue2021transporting} investigates on the underlying relationship between feature and labels. Another approach concerns the use of normalization, in particular Batch Normalization-based methods~\cite{chang2019domain,li2018adaptive,mirza2022norm,romijnders2019domain} which hypothesize that domain related knowledge is represented by the statistics of the Batch Normalization (BN)~\cite{ioffe2015batch}, thus transferring source model to the target domain could be achieved by modulating the statistics in the BN layer. Such approach is straightforward in implementation and requires minimal computational resource.

\subsection{Comparison with Image-UDA Methods}
\label{section:closed-methods:compare_image}
It can be observed that various VUDA methods are constructed with inspiration from the success of image-based UDA methods. However, given the multi-modality nature of videos, there are noticeable differences between image-based UDA and VUDA methods. Tab.~\ref{table:3-b-comapre_image} presents the comparison between VUDA and image-based UDA methods from the perspectives of the modality used, source of data, and more common techniques. It is noticed that VUDA not only differs from image-UDA over the modalities and source of data used for training and testing, they also differ in the common techniques used. Particularly, while there are plenty of reconstruction-based image UDA methods~\cite{ghifary2016deep,jhuo2012robust,murez2018image}, there is currently only one VUDA method which is reconstruction-based. This is owe to the difficulty of generating videos, which is much computational costly and challenging due to the need to generate a sequence of correlated images. Meanwhile, VUDA methods tend to leverage semantic-based approach given the rich semantic information embedded in the different modalities.

\begin{table*}[!ht]
\center
\caption{Comparison between VUDA and image-based UDA.}
\resizebox{.8\linewidth}{!}{\noindent
\begin{tabular}{m{.25\linewidth}|C{.35\linewidth}|E{.35\linewidth}}
\hline
\hline
Comparison & VUDA & Image-based UDA \\
\hline
Modality used & Temporal, spatial, and audio & Spatial\\
\hline
Source of data &
Mainly Video datasets: e.g., UCF101~\cite{soomro2012ucf101}, HMDB51~\cite{kuehne2011hmdb}, Kinetics~\cite{kay2017kinetics}
& Image datasets: e.g., MNIST~\cite{deng2012mnist}, Caltech~\cite{bansal2021transfer}, VisDA~\cite{peng2017visda}\\
\hline
More common techniques & Adversarial-based, Semantic-based, Composite & Adversarial-based, Discrepancy-based, Reconstruction-based\\
\hline
\hline
\end{tabular}
}
\label{table:3-b-comapre_image}
\end{table*}

\begin{table*}[!t]
\center
\caption{Different categories of methods for non-closed-set VUDA.}
\resizebox{1.\linewidth}{!}{\noindent
\begin{tabular}{m{.23\linewidth}|C{.3\linewidth}|C{.25\linewidth}|E{.18\linewidth}}
    \hline
    \hline
    Differences from closed-set & Scenarios & Assumptions/Constraints & Methods \Tstrut\Bstrut\\
    \hline
    \multirow{2}{*}{Label space constraint\Tstrut}
    & \textit{partial-set} VUDA (PVDA) \Tstrut & $\mathcal{Y}_{T}\subset\mathcal{Y}_{S}$ and $|\mathcal{C}_{T}|<|\mathcal{C}_{S}|$ \Tstrut & PATAN~\cite{xu2021partial}, MCAN~\cite{wang2022calibrating}\\
    & \textit{open-set} VUDA (OSVDA) \Bstrut & $\mathcal{Y}_{S}\subset\mathcal{Y}_{T}$ and $|\mathcal{C}_{S}|<|\mathcal{C}_{T}|$ \Bstrut & DMDA~\cite{wang2021dual}\Bstrut\\
    \hline
    \multirow{7}{*}{Source data assumption\Tstrut}
    & \textit{multi-source} VUDA (MSVDA) \Tstrut & $M_{S}>1, M_{T}=1$ \Tstrut & TAMAN~\cite{xu2021multi}\\
    & \textit{source-free} VUDA (SFVDA) \Tstrut & $\mathcal{D}_{S}$ not accessible \Tstrut & ATCoN~\cite{xu2022learning}\\
    & \textit{black-box} VUDA (BVDA) \Tstrut & $\mathcal{D}_{S}$ and $\theta_{S}$ not accessible\Tstrut & EXTERN~\cite{xu2022extern}\\
    & \textit{test-time} VUDA (VTTA) \Tstrut & $\mathcal{D}_{S}$ not accessible, $\mathcal{D}_{T}$ accessed in online manner\Tstrut & MCTTA~\cite{zeng2023exploring}\\
    & Unsupervised Image-to-Video (UI2V) \Tstrut & Data in $\mathcal{D}_{S}$ are images\Tstrut & HPDA~\cite{chen2021spatial}, CycDA~\cite{lin2022cycda}\\
    \hline
    Target data assumption \Tstrut\Bstrut & \textit{zero-shot} VUDA (VDG) & $\mathcal{D}_{T}$ not accessible \Tstrut\Bstrut & VideoDG~\cite{yao2021videodg}, RNA-Net~\cite{planamente2022domain}\Tstrut\Bstrut\\
    \hline
    \multirow{7}{*}{Cross-domain tasks\Tstrut}
    & Temporal action segmentation\Tstrut & - & MTDA~\cite{chen2020mtda}, SSTDA~\cite{chen2020sstda}\\
    & Video semantic segmentation\Tstrut & - & DA-VSN~\cite{guan2021domain}, TPS~\cite{xing2022domain}\\
    & Video quality assessment\Tstrut & - & UCDA~\cite{chen2021unsupervised}\\
    & Video sign language recognition\Tstrut & - & Li~et. al~\cite{li2020transferring}\\
    & Video object tracking\Tstrut & - & Azimi~et. al~\cite{azimi2022self}\\
    \hline
    \hline
\end{tabular}
}
\label{table:4-closed_set-methods}
\end{table*}

\section{Methods for Non-Closed-Set Video Unsupervised Domain Adaptation}
\label{section:other-methods}

\begin{figure*}[ht]
  \centering
  \includegraphics[width=.9\linewidth]{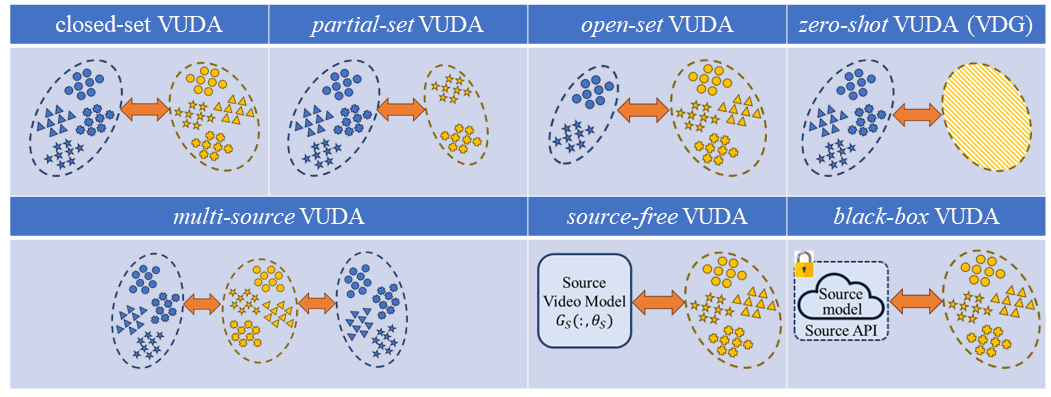}
  \caption{Comparing non-closed-set VUDA versus closed-set VUDA.}
  \label{figure:3-b-closed_vs_nonclosed}
\end{figure*}

The methods presented in Section~\ref{section:closed-methods} improve video model generalizability and enables knowledge to be transferred from a labeled source domain to an unlabeled target domain in the closed-set scenario. However, the constraints and assumptions of the closed-set VUDA may not hold in real-world scenarios, which could prompt concerns about model portability. In this section, we review deep learning-based VUDA methods in VUDA scenarios under different constraints and assumptions, categorized into four categories as introduced in Section~\ref{section:intro}. We first compare the different non-closed-set scenarios against the closed-set scenario, as presented in Fig.~\ref{figure:3-b-closed_vs_nonclosed}. We further summarize and compare all reviewed non-closed-set VUDA methods as presented in Tab.~\ref{table:4-closed_set-methods}.

\subsection{Methods with Differed Label Space Constraint}
\label{section:other-methods:label-space}
In the closed-set VUDA scenario, we assume that the source and target domain videos share the same label space (i.e.,~$\mathcal{Y}_{S}=\mathcal{Y}_{T}$ and $|\mathcal{C}_{S}|=|\mathcal{C}_{T}|$). With the presence of large-scale labeled public video datasets (e.g., Kinetics~\cite{kay2017kinetics} and Something-Something~\cite{goyal2017something}), it is more feasible in real-world scenarios to transfer representations learned in these large-scale video datasets to unlabeled small-scale video datasets. It is reasonable to assume that large-scale public video datasets can subsume categories of small-scale target datasets. Such a scenario is defined as the \textit{partial-set} VUDA, or the \textit{Partial Video Domain Adaptation} (PVDA)~\cite{xu2021partial}. It relaxes the constraint of identical source and target label spaces by assuming that the target label space is a subspace of the source one (i.e., ~$\mathcal{Y}_{T}\subset\mathcal{Y}_{S}$ and $|\mathcal{C}_{T}|<|\mathcal{C}_{S}|$). Compared to the closed-set VUDA, tackling PVDA poses more challenges due to the existence of outlier label space in the source domain denoted as $\mathcal{Y}_{out} = \mathcal{Y}_{S}\backslash \mathcal{Y}_{T}$, which causes negative transfer effect to the network's performance on the target domain. Meanwhile, during the training of the network, only labels of the target domain data are unknown, hence the part of which $\mathcal{Y}_{S}$ shares with $\mathcal{Y}_{T}$ is unknown.

With the inclusion of temporal features and multi-modal information (e.g., optical flow or audio), PVDA is also more challenging than its image-based counterpart (PDA~\cite{cao2018partial}) as negative transfer could be additionally triggered by the alignment of either temporal features or the multi-modal information. The key to tackling PVDA lies in mitigating negative effects brought by the unknown outlier label space $\mathcal{Y}_{S}\backslash \mathcal{Y}_{T}$ leveraging on the additional temporal or multi-modality features effectively. The pioneering work, Partial Adversarial Temporal Attentive Network (PATAN)~\cite{xu2021partial} proposes to tackle PVDA by the filtering of source-only outlier classes to mitigate negative transfer. To achieve this, PATAN leverages temporal features from two perspectives. Firstly, PATAN constructs temporal features such that those in outlier source-only classes discriminate those in the target classes by an attentive combination of local temporal features, where the attention builds upon the contribution of the local temporal features towards the class filtration process where source-only classes are filtered. Secondly, PATAN exploits temporal features toward the filtration of source-only classes to alleviate the effects of the misclassification of spatial features.

Later, the Multi-modality Cluster-calibrated partial Adversarial Network (MCAN)~\cite{wang2022calibrating} also tackles PVDA by filtering source-only outlier classes, exploiting the multi-modal features (optical flow feature) in addition to the spatial and temporal features leveraged in PATAN. MCAN further improves PVDA performance by dealing with label distribution shift~\cite{garg2020unified} across the source and target domain by clustering video features and weighing them accordingly such that MCAN promotes positive transfers of relevant source data and suppresses negative transfers of irrelevant source data jointly. Meanwhile, importance weighting~\cite{zhang2018importance} is an alternative approach for PVDA which leverages a two domain classifier strategy, where one domain classifier assigns the importance score of source samples and the other is applied to reduce the Jensen-Shannon divergence between the weighted source and target density.

Another practical VUDA scenario with differed label space assumption takes agnostic classes into consideration, assuming that there are unknown action classes in the target video domain, denoted as the \textit{open-set} VUDA, or the \textit{Open-Set Video Domain Adaptation} (OSVDA). Under such scenario, the source label space is a subspace of the target one (i.e.,,~$\mathcal{Y}_{S}\subset\mathcal{Y}_{T}$ and $|\mathcal{C}_{S}|<|\mathcal{C}_{T}|$). To tackle OSVDA, the Dual Metric Domain Adaptation framework (DMDA)~\cite{wang2021dual} is proposed which involves spatial and temporal features. DMDA deals with OSVDA by a Dual Metric Discriminator (DMD) which measures similarities between source and target video samples with a pre-trained classifier combined with prototypical optimal transport, applied to the frame, clip, and video levels. The DMD is further exploited as an initial separation and trains a binary discriminator to further distinguish whether target samples belong to the source action classes or the agnostic action classes.
Lately, AutoLabel~\cite{zara2023autolabel} is proposed as the first approach to tackle OSVDA by harnessing semantic information via CLIP~\cite{radford2021learning} which is a powerful Large Language-Vision Model. The harnessed semantic information is then leveraged to filter out and discover target-private classes names through an automatic labelling framework which incorporates an off-the-shelf Transformer-based~\cite{vaswani2017attention} image captioning network ViLT~\cite{kim2021vilt} applied towards each video frame. The discovered target-private classes names are then utilized to promote better separation of shared and target-private instances for more effective OSVDA.

\subsection{Methods with Differed Source Data Assumption}
\label{section:other-methods:source-data}
In addition to the constraints of the same label space across the source and target video domains, closed-set VUDA also makes several assumptions about the source data. Specifically, closed-set VUDA first assumes that there would be only $M_{S}=1$ video source domain with videos matching a uniform data distribution whose knowledge would be transferred to the target domain. In practice, source data are more likely to be collected from multiple datasets (e.g., the action ``Diving" can be found in datasets UCF101~\cite{soomro2012ucf101}, Kinetics~\cite{kay2017kinetics} and Sports-1M~\cite{karpathy2014large}). This VUDA scenario is defined as the \textit{multi-source} VUDA, or the \textit{Multi-Source Video Domain Adaptation} (MSVDA)~\cite{xu2021multi} which relaxes the constraint of identical source video data distribution by assuming that source video data are sampled from $M_{S}>1$ video domains corresponding to different video data distributions. The challenges of MSVDA lie in the negative transfer that would be triggered if domain shifts between multiple domain pairs are reduced directly regardless of their inconsistencies caused by distinct spatial and temporal feature distributions. The Temporal Attentive Moment Alignment Network (TAMAN)~\cite{xu2021multi} is a discrepancy-based method designed for MSVDA. It deals with MSVDA by constructing global temporal features via attentively combining local temporal features, where the attention strategies depend on both the local temporal feature classification confidence, as well as the disparity between the global and local feature discrepancies. Furthermore, TAMAN aligns spatial-temporal features jointly by aligning the moments of both spatial and temporal features across all domain pairs.

Closed-set VUDA also assumes that the source domain videos $V_{S}\in\mathcal{D}_{S}$ are always accessible during adaptation. However, action information in the source video domain usually contains the private and sensitive information of the actors, including their actions and the relevant scenes which is usually irrelevant to those in the target domain in real-world applications and should be protected from the target domain. For example, in hospitals, the anomaly action recognition of patients is usually required but videos that contain patients’ information cannot be shared across different hospitals. Current closed-set VUDA methods would therefore raise serious privacy and model portability issues, which are more severe than that raised by image-based domain adaptation. To address the video data privacy and model portability issue, a more practical VUDA scenario is formulated as the \textit{source-free} VUDA, or \textit{Source-Free Video Domain Adaptation} (SFVDA). In this VUDA scenario, only the well-trained source video models denoted as $G_{S}(:, \theta_{S})$, would be provided along with the unlabeled target video domain data for adaptation. Here $\theta_{S}$ is the parameter of $G_{S}$. With the absence of source data, the VUDA methods as reviewed that require data from both target and source domains for implicit or explicit alignment cannot be applied.

Recently, the Attentive Temporal Consistent Network (ATCoN)~\cite{xu2022learning} is proposed to deal with SFVDA. ATCoN aims to tackle SFVDA by obtaining temporal features that satisfy the cross-temporal hypothesis which hypothesizes that local temporal features (clip features) are not only discriminative but also consistent across each other and possess similar feature distribution patterns. This hypothesis is satisfied by ATCoN through learning temporal consistency composed of both feature and source prediction consistency. ATCoN further aligns target data to the source data distribution without source data access by attending to local temporal features with high source prediction confidence.
Subsequently, MTRAN~\cite{huang2022relative} attempts to tackle SFVDA by explicitly imitating and diminishing the video domain shifts along both the multimodal and temporal aspects by synthesizing hybrid samples based on source-like and target-like videos and aligning the representation of each synthetic sample towards the source distribution. Further,  STHC~\cite{li2023source} is proposed to perform stochastic spatio-temporal augmentations on each video and enforces prediction consistency from spatial, temporal, and historical perspectives for SFVDA. Very recently, DALL-V~\cite{zara2023dallv} attempts to leverage CLIP~\cite{radford2021learning} as a backbone Large Language-Vision Model and distill knowledge from CLIP for SFVDA. Instead of training the entire source model, DALL-V tackles SFVDA with only source adapters which are then learned by a student target adapter, where adapters are minimal networks aiming to preserve the generability of the large CLIP backbone while tackling the SFVDA scenario.

While SFVDA attempts to address privacy concerns in VUDA, it still relies on the well-trained source model parameters which allow generative models~\cite{goodfellow2014generative} to recover source videos. Inspired by black-box unsupervised domain adaptation~\cite{yang2022divide} in image-based domain adaptation, the \textit{black-box} VUDA or \textit{Black-box Video Domain Adaptation} (BVDA) is formulated more recently where the source video model is provided for adaptation only as a black-box predictor (e.g.,~API service). In other words, both the source domain $\mathcal{D}_{S}$ and $\theta_{S}$ are not accessible. To tackle the more challenging BVDA, EXTERN~\cite{xu2022extern} is proposed which is designed to adapt target models to the embedded semantic information of the source data resorting to the hard or soft predictions of the target domain from the black-box source predictor. EXTERN aims to extract effective temporal features in a self-supervised manner with high discriminability and complies with the cluster assumption~\cite{rigollet2007generalization} where regularizations are applied over clip features.

A further variant of the SFVDA scenario concerns the fact that test domain data may not be accessed as a whole, but rather accessed in small batches where the target model is optimized by finetuning the source model with the different batches of data in an online manner. Such scenario is termed as the\textit{test-time} VUDA or \textit{Video Test-time Adaptation} (VTTA). Since target data can only be accessed in a batch-wise manner, prior methods for SFVDA~\cite{xu2022learning,xu2022extern} that involves the overall distribution of target data are not applicable. MCTTA~\cite{zeng2023exploring} tackles VTTA by constructing an auxiliary motion encoding module which is a duplication of deep layers of the trained source model. Fast and slow clips of the target video are sampled separately, which are leveraged by MCTTA to perform the fast-to-slow unidirectional alignment and calculate classification loss based on stochastic pseudo-negative sampling. These fast and slow clips are also encoded by a momentum encoder for slow-fast dual contrastive learning. 

Apart from the difference in the number of source domains and the accessibility of source domain data, one other variant of closed-set VUDA assumes that the source domain data type differs from that of the target domain. Specifically, since it is more likely and feasible to gather labeled image data with less storage need, it is worth exploring if knowledge obtained from images, could aid the understanding of videos. Such cross-domain task thus assumes that the source domain data are still images, forming the Unsupervised Image-to-Video (UI2V) task. To effectively transfer knowledge learnt from still images that tend to focus on representative moments towards videos, proposed methods need to overcome the significant modality difference in both the representation space and the physical meaning of data. One primary work, the IVA~\cite{zhang2016semi} presented a classifier based image-to-video adaptation method and maps the common features of images and videos by kernel PCA~\cite{scholkopf1998nonlinear}. Subsequently, DIVAFN~\cite{liu2019deep} presented a  unified learning framework with individual neural networks for images and videos, where the modality shift among images and videos are reduced through a  cross-modal similarity. DIVAFN further leverage the semantic information as a guide to fuse the image and video features, while fusing their representation through semantic autoencoders. More recently, HPDA~\cite{chen2021spatial} tackles UI2V by a a spatial-temporal causal inference framework where a spatial-temporal causal graph is built to infer the effects of the spatial and temporal domain shifts. HPDA adaptively reduce the spatial and temporal domain shifts under the guidance of counterfactual causality via a causality-guided bidirectional heterogeneous mapping. Class-wise alignment is also leveraged in HPDA to enhance the ability to exploit images for video recognition. Meanwhile, CycDA~\cite{lin2022cycda} proposes to decouple the domain-alignment and spatio-temporal learning to bridge the modality gap between images and videos. CycDA further proposes a cyclic alternation between spatial and spatio-temporal learning to improve spatial and spatio-temporal models respectively by passing pseudo-labels learned on both models to supervise each other in a cycle. ST-I2V~\cite{zhuo2023synthesizing} is proposed as a single-stage method that synthesizes video from the source static image and convert the image-to-video adaptation problem into video-to-video adaptation problem. ST-I2V encourages the classification responses on target videos to maintain discriminative and diversity. ST-I2V further proposes a new pseudo label generation method that inherit the robustness of class-wise prototypes.

\subsection{Methods with Differed Target Data Assumption}
\label{section:other-methods:target-data}
Besides assumptions made on the source data, closed-set VUDA also supposes that the target domain data is readily available. This assumption may also not hold when applying to actual applications, since prior knowledge of target data distribution is not guaranteed. It is more practical to assume that the target domain is unseen (i.e., data of the target domain is unavailable) during adaptation, defined as the \textit{zero-shot} VUDA, or \textit{Video Domain Generalization} (VDG). Closed-set VUDA methods are also inadequate towards VDG, since similar to SFVDA, the domain discrepancy between source and target domains cannot be computed or estimated without knowledge of target data distribution. VideoDG~\cite{yao2021videodg} identifies the key towards VDG is to strike a balance between generalizability and discriminability, achieved by expanding the frame relations of the source domain such that they are diverse enough to be generalized to potential target domains while remaining discriminative. Inspired by the Transformer~\cite{vaswani2017attention} and the Adversarial Domain Augmentation (ADA)~\cite{volpi2018generalizing}, VideoDG reaches such balance with the introduction of the Adversarial Pyramid Network (APN) trained with Robust Adversarial Domain Augmentation (RADA). Meanwhile, the RNA-Net~\cite{planamente2022domain} makes use of the multi-modal nature of videos by leveraging both audio and RGB features for tackling VDG. RNA-Net suggests that fusing multi-modal information naively may not bring improvements to the generalizability of models~\cite{wang2020makes} due to certain modalities being privileged over others. Therefore, it proposes a cross-modal audio-visual Relative Norm Alignment (RNA) loss which aims to progressively align the relative feature norms of the two modalities from the source domains such that domain-invariant audio-visual features are obtained for VDG.

Besides these VDG works which are generally considered as performing task-based adaptation (i.e., a single action class which is considered as a single task) on short video clips, there have been a few recent works that performs step-based adaptation on longer videos, in particular instructional videos which contains several steps for each video. Zhukov et al.~\cite{zhukov2019cross} proposed a component model that shares information between steps, and a weakly supervised learning framework that can handle the component model together with constraints incorporating different forms of weak supervision. Meanwhile, Li et al.~\cite{li2022gain} proposed a causal inference approach to cut off the excessive contextual dependency for enhancing video model generalizability for improved VDG performance.

\subsection{Methods with Differed Cross-Domain Tasks}
\label{section:other-methods:task}
For all the aforementioned works, the VUDA methods are designed for the cross-domain action recognition task, which is one of the most fundamental video-based task~\cite{herath2017going,poppe2010survey}. Besides action recognition, there have been various studies on other cross-domain video tasks, such as cross-domain temporal action segmentation. For the cross-domain temporal action segmentation task, the Mixed Temporal Domain Adaptation (MTDA)~\cite{chen2020mtda} is proposed as an adversarial-based method. MTDA deals with cross-domain temporal action segmentation by jointly aligning local and global embedded feature spaces while integrating a domain attention mechanism based on domain predictions to aggregate domain-specific frames for constructing global video representations. Subsequently, the Self-Supervised Temporal Domain Adaptation (SSTDA)~\cite{chen2020sstda} is proposed for the same cross-domain temporal action segmentation task. SSTDA leverages the adversarial-based approach for aligning source and target videos by integrating two self-supervised auxiliary tasks (i.e., the binary and sequential domain prediction tasks), performed on the frame and clip levels respectively.

Cross-domain video semantic segmentation is another task studied for VUDA, which is relevant towards robust and efficient deployment of vision-based autonomous driving systems~\cite{huang2019apolloscape,siam2021video}. DA-VSN~\cite{guan2021domain} is one pioneer work that tackles cross-domain video semantic segmentation by introducing temporal consistency regularization (TCR) to bridge the domain gap. The TCR consists of two components, the cross-domain TCR which minimizes the discrepancy between source and target video domains by guiding target predictions to have the same temporal consistency as the source ones, and the intra-domain TCR which guides unconfident target predictions to have the same temporal consistency as the confident ones. The Temporal Pseudo-Supervision (TPS)~\cite{xing2022domain} is next proposed inspired by the success of consistency training~\cite{xie2020unsupervised,ouali2020semi,melas2021pixmatch} on image-based domain adaptation. TPS explores consistency training in the spatiotemporal feature space by enforcing model predictions to be invariant to cross-frame augmentations which are applied to the unlabeled target video frames.
More recently, STPL~\cite{lo2023spatio} explores the cross-domain video semantic segmentation task under the source-free setting. The proposed STPL framework is a contrastive learning framework which consists of two main stages for spatio-temporal feature extraction and pixel-level contrastive learning. STPL tackles the source-free cross-domain video semantic segmentation by optimizing the pixel-level contrastive loss between the original and augmented spatio-temporal features.

Besides the aforementioned segmentation tasks, an Unsupervised Curriculum Domain Adaptation (UCDA)~\cite{chen2021unsupervised} is proposed for the task of cross-modal video quality assessment (VQA) with VUDA which aims to output the quality score of the target videos given a set of source videos. UCDA deals with the task of cross-domain VQA through a two-stage adversarial adaptation with an uncertainty-based ranking function to sort the samples from the target domain into a different subdomain. Cross-domain video sign language recognition~\cite{li2020transferring} is another novel task investigated by VUDA research that aims to recognize isolated sign words from the sign words available in web news. Li~et. al~\cite{li2020transferring} propose a coarse domain alignment approach for this task by jointly training a classifier on news signs and isolated signs to reduce the domain gap. In addition, they develop a prototypical memory to learn a domain-invariant descriptor for each isolated sign.
Besides, the video object tracking task~\cite{yao2020video} has also been evaluated across different domains, specifically in the VTTA scenario. Azimi et. al.~\cite{azimi2022self} evaluates how self-supervised methods leveraging correspondence matching based on the colorizing of frames~\cite{jabri2020space,lai2020mast} behave in the VTTA scenario by combining the self-supervised objectives with image-based TTA methods such as TENT~\cite{wang2020tent} and TTT~\cite{sun2020ttt}.
\begin{table*}[!htbp]
\center
\caption{Comparison of current cross-domain video benchmark datasets.}
\resizebox{1.\linewidth}{!}{
\begin{tabular}{l|ccccccc}
\hline
\hline
Dataset & \# Classes & \# Train/Test Videos & Source of Data & VUDA Scenarios & Tasks & Year & Website \\
\hline
UCF-Olympic~\cite{sultani2014human} & 6 & 851/294 & UCF50, Olympic Sports & Closed-Set & Action recognition & 2014 & \href{https://github.com/cmhungsteve/TA3N}{Website}\Tstrut\\[0.05cm]
UCF-HMDB\textsubscript{small}~\cite{sultani2014human} & 5 & 832/339 & UCF50, HMDB51 & Closed-Set & Action recognition & 2014 & \href{https://github.com/cmhungsteve/TA3N}{Website} \\[0.05cm]
UCF-HMDB\textsubscript{full}~\cite{chen2019temporal} & 12 & 2278/931 & UCF50, HMDB51 & Closed-Set & Action recognition & 2019 & \href{https://github.com/cmhungsteve/TA3N}{Website} \\[0.05cm]
Kinetics-Gameplay~\cite{chen2019temporal} & 12 & 46003/3995 & Kinetics-600, Gameplay & Closed-Set & Action recognition & 2019 & \href{https://github.com/cmhungsteve/TA3N}{Website} \\[0.05cm]
HMDB-ARID~\cite{xu2021arid} & 11 & 3058/1153 & HMDB51, ARID & Closed-Set & Action recognition & 2021 & \href{https://xuyu0010.github.io/vuda.html}{Website} \\[0.05cm]
Kinetics$\to$NEC-Drone~\cite{choi2020necdrone} & 7 & Total 5250 & Kinetics-400, NEC-Drone & Closed-Set & Action recognition & 2020 & \href{https://www.nec-labs.com/~mas/NEC-Drone/}{Website} \\[0.05cm]
Mixamo$\to$Kinetics~\cite{da2022dual} & 14 & 24533/11662 & Mixamo, Kinetics-700 & Closed-Set & Action recognition & 2022 & \href{https://github.com/vturrisi/CO2A}{Website} \\[0.05cm]
ActorShift~\cite{zhang2022audio} & 7 & 1305/200 & Kinetics-700, YouTube & Closed-Set & Action recognition & 2022 & \href{https://xiaobai1217.github.io/DomainAdaptation/}{Website} \\[0.05cm]
Kinetics$\to$BABEL~\cite{lee2024glad} & 12 & 17844/1102 & \centering Kinetics-400, BABEL & Closed-Set & Action Recognition & 2024 & \href{https://github.com/KHU-VLL/GLAD}{Website}\Bstrut\\[0.05cm]
UCF-HMDB\textsubscript{\textit{partial}}~\cite{xu2021partial} & 14 & 2304/476 & UCF101, HMDB51 & Partial & Action recognition & 2021 & \href{https://xuyu0010.github.io/pvda.html}{Website} \\[0.05cm]
MiniKinetics-UCF~\cite{xu2021partial} & 45 & 20996/1106 & MiniKinetics, UCF101 & Partial & Action recognition & 2021 & \href{https://xuyu0010.github.io/pvda.html}{Website} \\[0.05cm]
HMDB-ARID\textsubscript{\textit{partial}}~\cite{xu2021partial} & 10 & 2712/540 & HMDB51, ARID & Partial & Action recognition & 2021 & \href{https://xuyu0010.github.io/pvda.html}{Website} \\[0.05cm]
EPIC Kitchens~\cite{munro2020multi} & 8 & 7935/2159 & EPIC Kitchens & Closed-Set & Action recognition & 2020 & \href{https://EPIC Kitchens.github.io/2021}{Website} \\[0.1cm]
Daily-DA~\cite{xu2021multi} & 8 & 16295/2654 & \parbox{0.2\linewidth}{\centering ARID, HMDB51, Moments in Time, Kinetics-600} & \parbox{0.15\linewidth}{\centering Multi-Source/ Closed-Set} & Action recognition & 2021 & \href{https://xuyu0010.github.io/msvda.html}{Website}\Tstrut\Bstrut\\[0.45cm]
Sports-DA~\cite{xu2021multi} & 23 & 36003/4712 & \parbox{0.2\linewidth}{\centering UCF101, Sports-1M, Kinetics-600} & \parbox{0.15\linewidth}{\centering Multi-Source/ Closed-Set} & Action recognition & 2021 & \href{https://xuyu0010.github.io/msvda.html}{Website}\Tstrut\Bstrut\\[0.2cm]
\parbox{0.25\linewidth}{Something-Something~\cite{yao2021videodg}} & 20 & 9530/4000 & \parbox{0.2\linewidth}{\centering Something-Something} & Zero-Shot & Action Recognition & 2019 & \href{https://github.com/thuml/VideoDG}{Website}\Bstrut\\[0.05cm]
CrossTask~\cite{zhukov2019cross} & 83 & 1950/2763 & \parbox{0.2\linewidth}{\centering -} & Zero-Shot & Action Recognition & 2019 & \href{https://github.com/DmZhukov/CrossTask}{Website}\Bstrut\\[0.05cm]
\parbox{0.25\linewidth}{GAIN~\cite{li2022gain}} & 20 & Total 6382 & \parbox{0.2\linewidth}{\centering -} & Zero-Shot & Action Recognition & 2022 & \href{https://jun-long-li.github.io/GAIN/}{Website}\Bstrut\\[0.05cm]
\parbox{0.12\linewidth}{VIPER$\to$Cityscapes-Seq~\cite{guan2021domain}} & 30 & 136645/500 & VIPER, Cityscapes-Seq & Closed-Set & Semantic segmentation & 2021 & \href{https://github.com/Dayan-Guan/DA-VSN}{Website}\\[0.25cm]
\parbox{0.25\linewidth}{SYNTHIA-Seq$\to$Cityscapes-Seq~\cite{guan2021domain}} & 30 & 10975/500 & \parbox{0.2\linewidth}{\centering SYNTHIA-Seq, Cityscapes-Seq} & Closed-Set & Semantic segmentation & 2021 & \href{https://github.com/Dayan-Guan/DA-VSN}{Website}\Bstrut\\[0.25cm]
\parbox{0.25\linewidth}{CamVid-Sunny$\to$CamVid-Foggy~\cite{shin2021unsupervised}} & 32 & 577/124 & \parbox{0.2\linewidth}{\centering CamVid} & Closed-Set & Semantic segmentation & 2019 & \href{http://mi.eng.cam.ac.uk/research/projects/VideoRec/CamVid/}{Website}\Bstrut\\[0.25cm]
\hline
\hline
\end{tabular}
}
\label{table:5-1-benchmark}
\end{table*}

\begin{table*}[!t]
\center
\caption{Average accuracy ($\%$) on Primary VUDA Datasets. Methods are arranged in chronological order.}
\resizebox{.8\linewidth}{!}{
\begin{tabular}[b]{l|cc|ccc}
\hline
\hline
Methods & Backbones & Categories & UCF-Olympic & UCF-HMDB\textsubscript{\textit{small}} & UCF-HMDB\textsubscript{\textit{full}}\Tstrut\Bstrut\\
\hline
AMLS~\cite{jamal2018deep} & C3D~\cite{tran2015learning} & Discrepancy & 85.24 & 92.45 & -\Tstrut\\
DAAA~\cite{jamal2018deep} & C3D~\cite{tran2015learning} & Adversarial & 90.78 & - & -\\
TA\textsuperscript{3}N~\cite{chen2019temporal} & TRN~\cite{zhou2018temporal} & Adversarial & 95.54 & 99.40 & 80.06\\
TCoN~\cite{pan2020adversarial} & TRN~\cite{zhou2018temporal} & Adversarial & 94.95 & 96.78 & 88.15\\
VideoDG (VDG)~\cite{yao2021videodg} & TRN~\cite{zhou2018temporal} & - & - & - & 67.00\\
SAVA~\cite{choi2020shuffle} & I3D~\cite{carreira2017quo} & Composite & - & - & 86.70\\
MM-SADA~\cite{munro2020multi} & I3D~\cite{carreira2017quo} & Adversarial & - & - & 87.65\\
PASTN~\cite{gao2020pastn} & TR3D~\cite{gao2020pastn} & Composite & 99.05 & - & -\\
STCDA~\cite{song2021spatio} & I3D~\cite{carreira2017quo} & Semantic & 99.35 & 97.20 & 87.60\\
CMCo~\cite{kim2021learning} & I3D~\cite{carreira2017quo} & Semantic & - & - & 88.75\\
CoMix~\cite{sahoo2021contrast} & I3D~\cite{carreira2017quo} & Semantic & - & - & 90.30\\
MAN~\cite{gao2021novel} & ResNet-152~\cite{he2016deep} & Composite & 94.80 & - & -\\
ACAN~\cite{xu2021aligning} & MFNet~\cite{chen2018multi} & Composite & - & - & 89.50\\
CO\textsuperscript{2}A~\cite{da2022dual} & I3D~\cite{carreira2017quo} & Semantic & 98.75 & - & 91.80\\
MA\textsuperscript{2}l-TD~\cite{chen2022multi} & ResNet-101~\cite{he2016deep} & Adversarial & 97.36 & 99.40 & 85.80\\
CIA~\cite{yang2022interact} & I3D~\cite{carreira2017quo} & Adversarial & - & - & 93.26\\
DVM~\cite{wu2022dynamic} & TSM~\cite{lin2020tsm} & Semantic & 96.37 & - & 92.77\\
TranSVAE~\cite{wei2022unsupervised} & I3D~\cite{carreira2017quo} & Reconstruction & - & - & 93.37\\
ATCoN (SFVDA)~\cite{xu2022learning} & TRN~\cite{zhou2018temporal} & - & - & - & 82.51\\
EXTERN (BVDA)~\cite{xu2022extern} & TRN~\cite{zhou2018temporal} & - & - & - & 90.42\\
\hline
\hline
\end{tabular}
}
\label{table:5-2-primary}
\end{table*}

\section{Benchmark Datasets for Video Unsupervised Domain Adaptation}
\label{section:datasets}

An important factor in the development of deep learning methods is the availability of relevant datasets for the training and evaluation of the proposed methods. This also applies to the development of research in VUDA methods. Over the past decade, there has been a significant increase in cross-domain video datasets, which greatly facilitates and promotes research in the various VUDA scenarios. In this section, we review and summarize existing cross-domain video datasets for VUDA.
An overall comparison of existing datasets over their major attributes (number of action classes, number of training/testing videos, source of data, etc.) is presented in Tab.~\ref{table:5-1-benchmark}. 
Furthermore, we show the average performance of the methods on their respective benchmarked datasets in Tabs.~\ref{table:5-2-primary},~\ref{table:5-3-larger_shift},~\ref{table:5-4-partial}~\ref{table:5-5-multi_domain},~and~\ref{table:5-6-segmentation}.
Note that due to the different backbones and training techniques applied by the different methods, direct comparison of their performance may not be fair and only serves as an intuitive reference towards the comparison of each method. All average performances are reported based on the original paper in which the respective methods are proposed when applicable.

\subsection{Primary VUDA Datasets}
\label{section:datasets:primary}
Earlier VUDA works typically rely on two sets of primary cross-domain action recognition datasets, namely the UCF-Olympic dataset~\cite{sultani2014human} and the UCF-HMDB dataset~\cite{sultani2014human}. The UCF-HMDB dataset~\cite{sultani2014human} is subsequently denoted as the UCF-HMDB\textsubscript{\textit{small}} dataset to differentiate with a later dataset. Specifically, the UCF-Olympic dataset is built across the UCF50~\cite{reddy2013recognizing} and the Olympic Sports~\cite{niebles2010modeling} datasets, while the UCF-HMDB\textsubscript{\textit{small}} dataset is built across the UCF50 and the HMDB51~\cite{kuehne2011hmdb} datasets. Both cross-domain datasets are of a very small scale with limited action classes and training/testing videos. Subsequently, a larger UCF-HMDB\textsubscript{\textit{full}}~\cite{chen2019temporal} dataset is introduced to facilitate further research on VUDA. The UCF-HMDB\textsubscript{\textit{full}} dataset is also built across the UCF50 and the HMDB51 datasets, but with more than doubled number of classes compared to the UCF-HMDB\textsubscript{\textit{small}} dataset and contains much more videos. The UCF-HMDB\textsubscript{full} dataset has become one of the most commonly used benchmark datasets for VUDA research.

\begin{table*}[!htbp]
\center
\caption{Average accuracy ($\%$) on VUDA datasets with larger domain shifts. Methods are arranged in chronological order.}
\resizebox{1.\linewidth}{!}{
\begin{tabular}[b]{l|cc|cccccc}
\hline
\hline
Methods & Backbones & Categories & Kinetics-Gameplay & HMDB-ARID & Kinetics$\to$NEC-Drone & Mixamo$\to$Kinetics & ActorShift & Kinetics$\to$BABEL\\
\hline
TA\textsuperscript{3}N~\cite{chen2019temporal} & TRN~\cite{zhou2018temporal} & Adversarial & 27.50 & 21.10 & 28.10 & 10.00 & - & -\\
NEC-Drone~\cite{choi2020necdrone} & I3D~\cite{carreira2017quo} & Composite & - & - & 15.10 & - & - & -\\
SAVA~\cite{choi2020shuffle} & I3D~\cite{carreira2017quo} & Composite & - & - & 31.60 & - & - & -\\
MM-SADA~\cite{munro2020multi} & SlowFast~\cite{feichtenhofer2019slowfast} & Adversarial & - & - & - & - & 62.60 & -\\
CoMix~\cite{sahoo2021contrast} & I3D~\cite{carreira2017quo} & Semantic & - & - & - & - & - & 21.40\\
ACAN~\cite{xu2021aligning} & MFNet~\cite{chen2018multi} & Composite & - & 52.20 & - & - & - & -\\
CO\textsuperscript{2}A~\cite{da2022dual} & I3D~\cite{carreira2017quo} & Semantic & - & - & 33.20 & 16.40 & - & 24.10\\
MA\textsuperscript{2}l-TD~\cite{chen2022multi} & ResNet-101~\cite{he2016deep} & Adversarial & 31.45 & - & - & - & - & -\\
A\textsuperscript{3}R~\cite{zhang2022audio} & SlowFast~\cite{feichtenhofer2019slowfast} & Semantic & - & - & - & - & 67.30 & -\\
GLAD~\cite{lee2024glad} & I3D~\cite{carreira2017quo} & Composite & - & - & - & - & - & 33.70\\
\hline
\hline
\end{tabular}
}
\label{table:5-3-larger_shift}
\end{table*}

\subsection{VUDA Datasets with Larger Domain Shifts}
\label{section:datasets:large-shift}
The aforementioned datasets are all built on datasets whose videos are collected mostly on web platforms (e.g., YouTube) with videos shot in normal conditions (e.g., normal illumination and contrast with clear pictures). Therefore the domain shifts across the different domains may not be significant. Consequently, the generalizability of VUDA approaches well-performed on the aforementioned datasets would be low in real-world applications where the domain shifts may be much larger. To cope with such limitations, cross-domain VUDA datasets with larger domain shifts are introduced. One example is the Kinetics-Gameplay~\cite{chen2019temporal} dataset which bridges real-world videos with virtual-world videos. Kinetics-Gameplay is built with the Kinetic~\cite{kay2017kinetics} and the Gameplay~\cite{chen2019temporal} datasets collected from current video games. Another cross-domain dataset that bridges real-world and synthetic videos is the Mixamo$\to$Kinetics dataset~\cite{da2022dual}, built as a uni-directional dataset to transfer knowledge from synthetic videos built from the Mixamo system to the real-world videos of the Kinetics dataset. Another scenario where large domain shifts may encounter during adaptation is between regular human-captured videos and drone-captured videos, where drone-captured videos possess unique characteristics thanks to their distinct motions and perspectives. The Kinetics$\to$NEC-Drone~\cite{choi2020necdrone} is introduced to leverage the existing large-scale Kinetics to aid video models to perform action recognition on the challenging drone-captured videos in the NEC-Drone~\cite{choi2020necdrone} dataset. Meanwhile, large domain shifts could also occur due to significant differences in video statistics, such as between videos shot under normal illumination and videos shot under low illumination (or more generally, between videos shot under normal environment and adverse environment). To explore how to leverage current datasets to boost performance on videos shot in adverse environments, the HMDB-ARID dataset~\cite{xu2021aligning} is introduced. This dataset comprises videos from the HMDB51 and the ARID~\cite{xu2021arid}, whose videos are shot under adverse illumination conditions and with low contrast. Lately, the ActorShift~\cite{zhang2022audio} dataset is proposed to research the domain shift between human and animal actions, which is the first dataset to consider non-human actions. The source domain of human actions is collected from Kinetics-700~\cite{smaira2020short} dataset while the target domain of animal actions is collected directly from YouTube with the relevant action classes.
Subsequently, the Kinetics$\to$BABEL dataset is presented to explore on leveraging real-world action videos for the understanding of synthetic action videos. Real-world videos are collected from Kinetics-400~\cite{kay2017kinetics} while the synthetic videos are collected from the BABEL dataset~\cite{punnakkal2021babel}. This dataset is challenging thanks to the significant appearance difference between real and synthetic videos, especially considering the fact that the BABEL dataset lacks background information, which results in additional background gaps between the two domains.

\begin{table*}[!htbp]
\center
\caption{Average accuracy ($\%$) on partial-set VUDA (PVDA) datasets. Methods are arranged in chronological order.}
\resizebox{.85\linewidth}{!}{
\begin{tabular}[b]{l|cc|ccc}
\hline
\hline
Methods & Backbones & Categories & UCF-HMDB\textsubscript{\textit{partial}} & MiniKinetics-UCF & HMDB-ARID\textsubscript{\textit{partial}} \\
\hline
TA\textsuperscript{3}N~\cite{chen2019temporal} & TRN~\cite{zhou2018temporal} & Adversarial & 60.59 & 61.97 & 21.25\\
SAVA~\cite{choi2020shuffle} & TRN~\cite{zhou2018temporal} & Composite & 65.93 & 66.58 & 23.72\\
PATAN (PVDA)~\cite{xu2021partial} & TRN~\cite{zhou2018temporal} & - & 81.83 & 76.04 & 30.54\\
MCAN (PVDA)~\cite{wang2022calibrating} & TSN~\cite{wang2016temporal} & - & 83.94 & 81.34 & 44.37\\
\hline
\hline
\end{tabular}
}
\label{table:5-4-partial}
\end{table*}

\subsection{Partial-set VUDA (PVDA) Datasets}
\label{section:datasets:pvda}
The datasets above are all constructed for the closed-set VUDA scenario where there exist only two source-target video domain pairs (i.e., Domain A$\to$Domain B and Domain B$\to$Domain A) whose label spaces are shared. However, as mentioned in Section~\ref{section:other-methods}, constraints and assumptions for closed-set VUDA may not hold in real-world scenarios. Therefore, other cross-domain video datasets are introduced to support and facilitate research on VUDA with different constraints and assumptions. For partial-set VUDA (PVDA), a collection of three cross-domain partial-set video datasets is introduced in~\cite{xu2021partial}, namely UCF-HMDB\textsubscript{\textit{partial}}, MiniKinetics-UCF, and HMDB-ARID\textsubscript{\textit{partial}}. Among which the UCF-HMDB\textsubscript{\textit{partial}} is constructed inspired by UCF-HMDB\textsubscript{\textit{full}}~\cite{chen2019temporal}, built across the UCF101 and HMDB51 datasets. The MiniKinetics-UCF dataset is of much larger scale (8$\times$ that of UCF-HMDB\textsubscript{\textit{partial}}) and is designed to validate the effectiveness of PVDA approaches on large-scale datasets. It is built from the Mini-Kinetics~\cite{xie2017rethinking} and UCF101 datasets. Meanwhile, the HMDB-ARID\textsubscript{\textit{partial}} is built inspired by the HMDB-ARID~\cite{xu2021aligning} dataset and aims to boost the performance of video models on low-illumination model leveraging normal videos under the partial-set VUDA with larger domain shift.

\begin{table*}[!htbp]
\center
\caption{Average accuracy ($\%$) on multi-domain VUDA datasets. Methods are arranged in chronological order.}
\resizebox{.8\linewidth}{!}{
\begin{tabular}[b]{l|cc|ccc}
\hline
\hline
Methods & Backbones & Categories & Epic-Kitchens & Daily-DA & Sports-DA \\
\hline
TA\textsuperscript{3}N~\cite{chen2019temporal} & TRN~\cite{zhou2018temporal} & Adversarial & 43.20 & 28.49 & 70.26\\
MM-SADA~\cite{munro2020multi} & I3D~\cite{carreira2017quo} & Adversarial & 50.30 & - & -\\
STCDA~\cite{song2021spatio} & I3D~\cite{carreira2017quo} & Semantic & 51.20 & - & -\\
CMCo~\cite{kim2021learning} & I3D~\cite{carreira2017quo} & Semantic & 51.00 & - & -\\
CoMix~\cite{sahoo2021contrast} & I3D~\cite{carreira2017quo} & Semantic & 43.20 & - & -\\
TAMAN (MSVDA)~\cite{xu2021multi} & TRN~\cite{zhou2018temporal} & - & - & 44.85 & 77.84\\
RNA-Net (VDG)~\cite{planamente2022domain} & I3D~\cite{carreira2017quo} & - & 51.06 & - & -\\
CIA~\cite{yang2022interact} & I3D~\cite{carreira2017quo} & Adversarial & 52.20 & - & -\\
A\textsuperscript{3}R~\cite{zhang2022audio} & SlowFast~\cite{feichtenhofer2019slowfast} & Semantic & 61.00 & - & -\\
TranSVAE~\cite{wei2022unsupervised} & I3D~\cite{carreira2017quo} & Reconstruction & 52.60 & - & -\\
ATCoN (SFVDA)~\cite{xu2022learning} & TRN~\cite{zhou2018temporal} & - & - & 33.53 & 73.85\\
EXTERN (BVDA)~\cite{xu2022extern} & TRN~\cite{zhou2018temporal} & - & - & 39.64 & 83.18\\
\hline
\hline
\end{tabular}
}
\label{table:5-5-multi_domain}
\end{table*}

\subsection{Multi-Domain VUDA Datasets}
\label{section:datasets:md-vuda}
There are also several more recent datasets that are more comprehensive that include multiple domains within the cross-domain dataset such that there are more than 2 possible source/target video domain pairs. For instance, the Epic-Kitchens~\cite{munro2020multi} cross-domain dataset contains 3 domains from videos of three different kitchens in the original Epic-Kitchens action recognition dataset~\cite{damen2018scaling} and therefore includes 6 different combinations of source/target video domain pairs. Note that we follow the literatures in~\cite{munro2020multi,choi2020shuffle,song2021spatio,sahoo2021contrast} and still refer the Epic-Kitchens cross-domain dataset as ``Epic-Kitchens". Epic-Kitchens is generally built from a single large-scale action recognition dataset and contains videos collected from a controlled environment. Subsequently, other multi-domain VUDA datasets are introduced that contain videos from a wider range of different scenes collected from various public datasets. Inspired by the success of DomainNet~\cite{peng2019moment} as a unified and comprehensive benchmark for evaluating image-based domain adaptation under both closed-set and multi-domain scenarios, the Sports-DA and Daily-DA cross-domain action recognition datasets~\cite{xu2021multi,xu2022learning} are introduced. The Daily-DA dataset contains 4 different domains constructed from HMDB51~\cite{kuehne2011hmdb}, ARID~\cite{xu2021arid}, Moment in Time~\cite{monfort2019moments}, and Kinetics-600~\cite{kay2017kinetics}, resulting in 12 different combinations of source/target video domain pairs, which is the largest number of source/target domain pairs to date. The Sports-DA dataset contains 3 different domains with sports videos from UCF101~\cite{soomro2012ucf101}, Kinetics-600~\cite{kay2017kinetics}, and Sports-1M~\cite{karpathy2014large}, resulting in 6 different combinations of source/target video domain pairs, and contains more action classes than that in both Epic-Kitchens and Daily-DA.

\subsection{Video Domain Generalization (VDG) Datasets}
\label{section:datasets:vdg}
The aforementioned datasets could also be leveraged to benchmark zero-shot VUDA (or VDG) methods, where UCF-HMDB\textsubscript{full} has been used to evaluate VideoDG~\cite{yao2021videodg} while Epic-Kitchens has been used to evaluate RNA-Net~\cite{planamente2022domain}, as presented in Tab.~\ref{table:5-2-primary} and~\ref{table:5-5-multi_domain}. Meanwhile, test videos may have similar but different consequences of actions in many real-world scenarios. The temporal domain shift is significant under such circumstances, thus requiring the benchmark to emphasize on temporal information. Yao et al.~\cite{yao2021videodg} therefore proposed a VDG dataset constructed upon the Something-something dataset~\cite{goyal2017something} by selecting 20 basic categories. Each category is divided into two sub-categories with different consequences of actions, such as doing something vs. pretending to do something. The final dataset contains 9,530 source videos and around 4,000 target videos.

Meanwhile, as mentioned in Section~\ref{section:other-methods:target-data}, there have been a few works which explore VDG in longer instructional videos through step-based adaptation. Two datasets: CrossTask~\cite{zhukov2019cross} and GAIN~\cite{li2022gain} have been proposed to evaluate video model generalizability and step-based adaptation performance. Both datasets contain instructional videos that are longer in length compared to previous datasets (e.g., Kinetics-600 or Moment in Time) and each contains multiple steps. Among which CrossTask~\cite{zhukov2019cross} includes 2,763 videos of 18 primary tasks that comprises 213 hours of video and act as the target domain, as well as 1.950 videos of 65 related tasks comprising 161 hours of video that act as the source domain. GAIN contains a total of 6,382 action segments in 1,231 instructional videos, which each video containing an average of 5 steps. The videos consist of 20 fine-grained action categories.

\begin{table*}[!htbp]
\center
\caption{Average IOU on VUDA datasets for cross-domain video semantic segmentation. Methods are arranged chronologically.}
\resizebox{.95\linewidth}{!}{
\begin{tabular}[b]{l|c|ccc}
\hline
\hline
Methods & Backbones & VIPER$\to$Cityscapes-Seq & SYNTHIA-Seq$\to$Cityscapes-Seq & CamVid-Sunny$\to$CamVid-Foggy\\
\hline
DA-VSN~\cite{guan2021domain} & ACCEL~\cite{jain2019accel} & 47.80 & 49.50 & - \\
VAT+VST~\cite{shin2021unsupervised} & ResNet-50~\cite{he2016deep}+DeepLabV2~\cite{chen2017deeplab} & 36.43 & - & 55.10 \\
TPS~\cite{xing2022domain} & ACCEL~\cite{jain2019accel} & 48.90 & 53.80 & - \\
\hline
\hline
\end{tabular}
}
\label{table:5-6-segmentation}
\end{table*}

\subsection{VUDA Dataset for Cross-Domain Video Semantic Segmentation}
\label{section:datasets:semantic}
While all aforementioned datasets are meant for the cross-domain action recognition task, research on VUDA is not limited to such a task as mentioned in Section~\ref{section:other-methods:task}. With increasing interest in other cross-domain video tasks, there have been some relevant datasets proposed. This is especially for cross-domain video semantic segmentation, where three cross-domain datasets are proposed: VIPER$\to$Cityscapes-Seq, SYNTHIA-Seq$\to$Cityscapes-Seq~\cite{guan2021domain}, and CamVid-Sunny$\to$CamVid-Foggy~\cite{shin2021unsupervised}, built from current semantic segmentation datasets. The VIPER$\to$Cityscapes-Seq is built from Cityscapes-Seq~\cite{cordts2016cityscapes} and VIPER~\cite{richter2017playing}, while the SYNTHIA-Seq$\to$Cityscapes-Seq from SYNTHIA-Seq~\cite{ros2016synthia} and Cityscapes-Seq. Both VIPER and SYNTHIA-Seq are synthetic videos generated from games or the Unity Development platform~\cite{haas2014unity} while Cityscapes-Seq is built with videos captured in real-world scenes. Meanwhile, CamVid-Sunny$\to$CamVid-Foggy~\cite{shin2021unsupervised} is built upon CamVid~\cite{brostow2009semantic}, and is constructed to evaluate the weather adaptation capability of cross-domain semantic segmentation methods. 

\begin{table*}[!htbp]
\center
\caption{Statistics of backbones leveraged for VUDA methods. Scores reported are formulated as best accuracy ($\%$) (difference in accuracy ($\%$)).}
\resizebox{.8\linewidth}{!}{
\begin{tabular}[b]{l|c|cccccc}
\hline
\hline
Backbones & Methods Used & UCF-Olympic & UCF-HMDB\textsubscript{\textit{small}} & UCF-HMDB\textsubscript{\textit{full}} & Epic-Kitchens & Daily-DA & Sports-DA \\
\hline
C3D~\cite{tran2015learning} & 2 & 90.78 (5.54) & 92.45 (-) & - (-) & - (-) & - (-) & - (-)\\
TRN~\cite{zhou2018temporal} & 6 & 95.54 (0.59) & 99.40 (2.62) & 90.42 (23.42) & 43.20 (-) & 44.85 (16.36) & 83.18 (12.92)\\
I3D~\cite{carreira2017quo} & 9 & 99.35 (0.60) & 97.20 (-) & 93.37 (6.67) & 52.60 (9.40) & - (-) & - (-)\\
TR3D~\cite{gao2020pastn} & 1 & 99.05 (-) & - (-) & - (-) & - (-) & - (-) & - (-)\\
ResNet-101~\cite{he2016deep} & 1 & 97.36 (-) & 99.40 (-) & 85.80 (-) & - (-) & - (-) & - (-)\\
ResNet-152~\cite{he2016deep} & 1 & 94.80 (-) & - (-) & - (-) & - (-) & - (-) & - (-)\\
MFNet~\cite{chen2018multi} & 1 & - (-) & - (-) & 89.50 (-) & - (-) & - (-) & - (-)\\
TSM~\cite{lin2020tsm} & 1 & 96.37 (-) & - (-) & 92.77 (-) & - (-) & - (-) & - (-)\\
SlowFast~\cite{feichtenhofer2019slowfast} & 1 & - (-) & - (-) & - (-) & 61.00 (-) & - (-) & - (-)\\
\hline
\hline
\end{tabular}
}
\label{table:5-a-comapre_backbone}
\end{table*}

\subsection{Backbones and Their Impact on VUDA Performances}
\label{section:datasets:backbone}
Though as mentioned in the beginning of this Section, direct comparison of the performances of the various VUDA methods is not fair due to the different backbones applied, there are still some interesting observations over the backbones chosen and how they impact on the final VUDA performances. As there are only a few results reported on VUDA datasets with larger domain shifts, PVDA datasets, VDG datasets and VUDA dataset for cross-domain video semantic segmentation, we focus our discussion only on primary VUDA datasets and multi-domain VUDA datasets. Tab.~\ref{table:5-a-comapre_backbone} compares the statistics of the backbones leverage, including the number of methods leveraging each backbone, as well as the best accuracy and the difference between the best and the worst accuracy reported.

While all the methods discussed leverage a CNN-based backbone, I3D~\cite{carreira2017quo} is by far the most popular backbone, with 9 out of 23 methods leverage it as the feature extractor backbone. This is thanks to its simple implementation with the availability of pre-trained models trained on Kinetics-400~\cite{carreira2017quo}. In addition, the computation cost of I3D is fair (compared to methods such as SlowFast~\cite{feichtenhofer2019slowfast}) with decent capability in modelling temporal features by inflating 2D-CNNs. We also observe the I3D could result in comparable performance for the zero-shot VUDA scenario compared to its closed-set VUDA performance, which demonstrate its generalizability. All these factors contribute to its wide application.

Apart from I3D, TRN~\cite{zhou2018temporal} is the second most popular backbone, thanks to its ability in obtaining effective temporal features through reasoning over correlations between spatial representations, which corresponds with how humans would recognize actions. Besides, compared to backbones such as I3D and C3D~\cite{tran2015learning}, TRN requires less input video frames, resulting in lower computation cost. However, it could be observe that the difference in accuracy is quite large between the different methods that leverage TRN, especially on UCF-HMDB\textsubscript{\textit{full}}. This large gap owes to the poor performance of VideoDG~\cite{yao2021videodg}, which suggest the poor generalizability of TRN.

While it is generally believed that a more complex network should result in generally better performance, current results does not support such hypothesis, in particular on primary VUDA datasets. For example, while ResNet-152~\cite{he2016deep} is significantly more complex than ResNet-101~\cite{he2016deep}, its performance is inferior on UCF-Olympics. However, it can be observed that networks that could obtain better temporal features (e.g., TSM and TRN) generally outperforms networks that are not specifically designed to accommodate temporal features (e.g., ResNet-101 and MFNet), demonstrating the importance of extracting and aligning temporal features for effective VUDA.
\section{Discussion: Recent Progress and Future Directions}
\label{section:discussion}

In this section, we summarize the recent progress in VUDA research with observations. We further analyze and provide our insights on possible future directions of development for VUDA research.

\subsection{Recent Progress in VUDA Research}
\label{section:discussion:progress}
Compared to earlier works, recent VUDA research has made significant progress from three perspectives: a) tackling VUDA under different scenarios; b) leveraging the multi-modality nature of videos; and c) exploiting shared semantics across domains with semantic-based methods.

\textbf{a) Tackling VUDA Under Different Scenarios.}
The closed-set scenario has been the focus of VUDA research thanks to its simplicity that results from the assumptions of a single pair of the labeled video source and unlabeled video target domains with the source videos and source models accessible, and the constraints of a shared label space across the source/target domain pair. However, as mentioned in Section~\ref{section:other-methods} and in~\cite{xu2021partial,xu2021multi,wang2022calibrating,xu2022learning}, closed-set VUDA may not be applicable in real-world scenarios. To cope with model portability and other (e.g., data privacy) issues caused by the constraints and assumptions of closed-set VUDA, several other scenarios of VUDA have been recently studied. These include the partial-set PVDA~\cite{xu2021partial,wang2022calibrating}, the open-set OSVDA~\cite{wang2021dual}, the multi-domain MSVDA~\cite{xu2021multi}, the SFVDA~\cite{xu2022learning} and BVDA~\cite{xu2022extern} with source-free/black-box source model settings, as well as the VDG~\cite{yao2021videodg} with target-free settings. The introduction of relevant datasets further promotes the research on various non-closed-set VUDA scenarios and further improves the capability of VUDA methods in real-world scenarios.

\textbf{b) Leveraging the Multi-Modality Nature of Videos.}
Tackling VUDA is more challenging than tackling image-based UDA largely thanks to the inclusion of the additional temporal features and features of other modalities (e.g., optical flow and audio) in videos. These additional features would all incur extra domain shifts across source and target domains. Earlier methods such as the adversarial-based TA\textsuperscript{3}N~\cite{chen2019temporal}, the discrepancy-based AMLS~\cite{jamal2018deep} or the composite VUDA method PASTN~\cite{gao2020pairwise} focus primarily on tackling domain shift caused by the additional temporal features. Subsequently, more recent methods such as MM-SADA~\cite{munro2020multi}, CIA~\cite{yang2022interact} and A\textsuperscript{3}R~\cite{zhang2022audio} have realized the importance of tackling domain shift caused by the features of different modalities, with tackling domain shifts from optical flow and audio features being the focus. Later methods have achieved notable improvements over the same benchmark against prior methods without multi-modal feature alignment, which proves the efficacy of aligning multi-modal features toward achieving effective VUDA. However, it should be noted that audio features may not be readily available in benchmark datasets or in real-world scenarios (e.g., surveillance footage or autonomous driving footage where the audio captured is mostly ambient noise). Therefore there are certain limitations in applying VUDA methods that exploit audio features.

\textbf{c) Exploiting Shared Semantics with Semantic-based VUDA Methods.}
Compared to adversarial-based and discrepancy-based VUDA methods, semantic-based VUDA methods have not been considered until more recently. This owes to the fact that aligning video domains by exploiting shared semantics is not as intuitive as aligning video domains by minimizing video domain discrepancies whether explicitly or implicitly. However, the performances of different semantic-based VUDA methods (e.g., CMCo~\cite{kim2021learning} and CoMix~\cite{sahoo2021contrast}) proves that exploiting shared semantics with spatio-temporal association, feature clustering and modality correspondence is beneficial towards obtaining domain-invariant video features. Comparatively, semantic-based methods are more stable in terms of optimization compared to adversarial-based methods, while obtaining superior performance than discrepancy-based methods. Furthermore, semantic-based VUDA methods can be combined with both adversarial and discrepancy-based VUDA methods to form composite methods such as SAVA~\cite{choi2020shuffle} and PASTN~\cite{gao2020pairwise}.

\subsection{Challenges of Current VUDA Research and its Future Directions}
\label{section:discussion:challenge-and-future}

Despite the notable progress made in VUDA research, there are still various challenges hampering the effectiveness of existing VUDA research. The challenges could generally be categorized into three categories: a) challenges in explored VUDA scenarios; b) challenges in the multi-modal information leveraged; and c) challenges in more effective VUDA methods with self-supervision. Dealing with these challenges would greatly benefit future VUDA methods and are considered as potential future directions for VUDA research.

\textbf{a) Challenges in Explored VUDA Scenarios, Settings and Tasks.}
There have been various non-closed-set VUDA scenarios researched as mentioned in Section~\ref{section:other-methods}. However, compared to domain adaptation in NLP and image tasks which has been researched more comprehensively~\cite{su2020active,prabhu2021active,peng2019federated}, we observe that there are still a number of scenarios that have not been touched upon for VUDA. For instance, while a method has been proposed for multi-source VUDA (MSVDA) where the constraint of $M_{S}=1$ video source domain is relaxed, the multi-target VUDA (MTVDA) where the constraint of $M_{T}=1$ video target domain is relaxed has not been touched upon in research. Combining active learning~\cite{ren2021survey} and VUDA that formulates active VUDA which aims to adapt the source video model to the target video domain by acquiring labels for a selected maximally-informative subset of target videos via an oracle is another scenario that has not been researched in VUDA. To further protect source data privacy, black-box VUDA is another feasible VUDA scenario where besides source video data, the source video model is also made inaccessible to the target video domain, which prevents source videos to be recovered by generative models~\cite{goodfellow2014generative}. Meanwhile, the data privacy concern is also applicable to the current MSVDA scenario where video data are accessible between different source domains. Combining federated learning~\cite{yang2019federated} with VUDA is a possible solution such that video data are not shareable between different video domains. The aforementioned VUDA scenarios are more realistic and future research on these scenarios could further boost the capability of VUDA methods in applying to real-world applications.

Apart from falling short of real-world scenarios, current VUDA approaches also fall behind other domain adaptation research in tackling different types of domain shift. Current VUDA approaches are all designed to tackle covariate shift~\cite{sugiyama2007covariate} between source and target domains, where the distribution of input changes but the conditional probability of the output given an input remains the same for source and target domains. Yet, the source and target domains could also differ in the distribution of the labels or the numerical output variables, resulting in label shift (also known as prior shift)~\cite{lipton2018detecting,tachet2020domain,wu2021online} across domains. Further, the conditional probability of the output given an input could also differ across source and target domains, causing conditional shift (also known as concept drift)~\cite{zhang2013domain,liu2021deep,liu2021adversarial} across domains. More recently, the causal conditional shift assumption~\cite{li2023transferable} has been discussed for time-series adaptation. Given the comprehensive causal interactions that exist in videos (e.g., causal interactions between objects shot and between the scenes shot in different frames)~\cite{ayazoglu2013finding,wang2022weakly}, the causal conditional shift assumption is also worth exploration for adaptation between different video domains.

Further, while there have been various VUDA studies on cross-domain video tasks other than action recognition (e.g., temporal action segmentation and video semantic segmentation), it is observed that these cross-domain video tasks are generally discriminative tasks. Meanwhile, there are still cross-domain video tasks, especially generative tasks, that have not been attempted. One example concerns the video prediction task~\cite{gao2019disentangling,oprea2020review}, where models are tasked to predict future frames given past video frames. With the increasing focus on data generation through deep learning models, extending VUDA studies towards cross-domain generative video tasks could further improve the applicability of VUDA approaches towards varied cross-domain video tasks.

\textbf{b) Challenges in Multi-Modal Information Leveraged.} 
The multi-modal information leveraged for existing VUDA methods is largely limited to RGB, optical flow, and audio information. However, videos also contain more modalities of information, which have already been exploited for supervised video tasks but not exploited for VUDA. A typical example involves the human skeleton data~\cite{ke2017new,song2017end,du2015hierarchical,duan2022revisiting} which is a compact and effective action descriptor that focuses on the temporal change of the pose of the actor and is known for its immune to contextual variation, such as background and illumination variation. Meanwhile, the input of existing VUDA methods rely only on RGB cameras, while videos could also be obtained by other sensors including depth cameras~\cite{bulbul2021gradient,xiao2019action}, infrared cameras~\cite{gao2016infar,liu2018global}, or even lidars~\cite{zhong2021human,you2022dynamic,benedek2016lidar}. Future VUDA methods should also take videos taken from these sensors into consideration.

Besides the challenges faced in leveraging the different categories of modalities, the performance of current VUDA methods may also be hampered by how currently leveraged modalities (especially the temporal modality) can be utilized effectively. The are mainly two factors to be concerned: how video features can be extracted effectively with temporal modality, and how temporal features can be better aligned. To extract video features, existing works still tend to leverage ResNet-based~\cite{he2016deep} CNN networks to obtain features from RGB, optical flow, and audio. More recent works~\cite{arnab2021vivit,liu2021swin,chen2021visformer,sui2022craft,gao2020learning,liu2022tcgl} have shown the efficacy of both Transformers-based~\cite{vaswani2017attention} and Graph Neural Networks (GNN)~\cite{scarselli2008graph,zhang2019heterogeneous} in obtaining effective features for downstream video tasks. It is intuitive that aligning video features effectively should be built on the assumption that the video features to align with are effective themselves. Lately, Transformer-based networks such as TimeSFormer~\cite{bertasius2021space} and GNN have been used in VUDA~\cite{xu2023augmenting,peng2023featfsda}, resulting in noticeable performance gain. Therefore, we foresee the trend of leveraging Transformer-based and GNNs in future VUDA methods as the backbone for effective feature extraction.

Despite the additional modalities contained in videos, most existing VUDA methods are developed from certain image-based UDA methods, while taking additional steps to align temporal features. These steps could be applying attention in the temporal dimension~\cite{chen2019temporal,xu2021partial} or applying semantic constraints in the temporal dimension through temporal consistency loss~\cite{xu2022learning,xu2022extern}. However, these methods are generally applied on trimmed video clips, and may not be applicable to longer videos which may be encountered in real-world scenarios. In the supervised setting, temporal features of such untrimmed long videos could be better obtained through fusing temporal clips sampled at different speed (e.g., SlowFast~\cite{feichtenhofer2019slowfast}) or by combining with temporal action proposal~\cite{lin2019bmn,escorcia2016daps} to extract meaningful actions temporally. We believe that aligning temporal features could be further explored inspired by these approaches.

\textbf{c) Challenges in VUDA Methods with Self-Supervision.} 
In recent years, there has been a significant increase in VUDA methods that leverage semantic-based or reconstruction-based approaches in full or in part, thanks to their high performance and extensibility towards combining with other approaches and towards different, more practical VUDA scenarios. Since target labels are unavailable for adaptation, both semantic-based and reconstruction-based approaches rely on self-supervision for obtaining shared cross-domain semantics or achieving data reconstruction. Among the various self-supervision techniques, contrastive learning has been widely utilized given the ease of formulation and their high performance. More recently, contrastive learning in visual tasks has been achieved not only by applying across visual features but also across visual features and their corresponding text labels or text descriptions~\cite{radford2021learning}. The strategy of learning visual concepts from natural language supervision in a contrastive manner results in more a generalizable network that could easily adapt to new visual tasks with only natural language cues from the pre-trained task. This is in line with the goal of VUDA which attempts to adapt networks to new video domains given information from the source domain, which includes text-based information such as video labels. Therefore, the exploration of self-supervised VUDA methods leveraging on natural language cues within the labeled source domain (e.g., source video labels) could be an interesting yet effective way towards further improvement of VUDA performances.

\textbf{d) Challenges in Benefiting from Large-Scale Pretrained LLVMs.}
Lately, a number of large pretrained models, especially Large Language-Vision Models (LLVMs) such as CLIP~\cite{radford2021learning}, FLAVA~\cite{singh2022flava}, have been introduced to the vision community. These pretrained LLVMs combines and embeds large-scale vision information with natural language information, and are equipped with strong zero-shot generalizability. They have benefited the video understanding community, especially for the supervised setting (with works e.g., ActionCLIP~\cite{wang2021actionclip} and FitCLIP~\cite{castro2022fitclip} for action recognition and CLIPBert~\cite{lei2021less} for text-to-video retrieval), and zero-shot setting with works such as Howto100M~\cite{miech2019howto100m}. Meanwhile, recent advancement has been made in image-based UDA where LLVMs are leveraged with prompt learning (e.g., DAPL~\cite{ge2022domain} and DPLCLIP~\cite{zhang2021domain}), resulting in significant UDA performance improvement with minimal training required thanks to the strong generalizability of pretrained LLVMs. Such advancement could also be extended to the video domains since prior works~\cite{wang2021actionclip,castro2022fitclip} demonstrates that the zero-shot generalizability of LLVMs holds for videos even with simple frame aggregation approaches such as temporal pooling or a single Transformer~\cite{vaswani2017attention} layer. However, it is noted that current LLVMs with prompt learning for image-based UDA also utilize explicit domain information (e.g., clip art or sketch), which may not be available for videos in either current cross-domain video benchmark datasets (as introduced in Section~\ref{section:datasets}) or in real-world scenarios.
There are indeed a few primary works that leveraged CLIP for VUDA, especially for SFVDA~\cite{zara2023dallv} and OSVDA~\cite{zara2023autolabel} scenarios where the generability and zero-shot capability of pre-trained CLIP boost the performances. However, these primary methods contain obvious limitations. Specifically, semantic information (e.g., classes names) obtained through CLIP only focus on modelling general objects and actors and cannot disambiguate among different actions involving the same objects and actors, while the filtering of semantic information requires significant involvement of expert knowledge. Meanwhile, pre-trained CLIP predictions may not be trustworthy and may elicit unpredictable behaviour during knowledge transfer. How future VUDA approaches could better benefit effectively from pretrained LLVMs subject to the different constraints in videos and improve their applicability in real-world tasks is well worth exploration.

\textbf{e) Challenges in Video-based Domain Adaptation Theory and VUDA Explainability.}
The development of domain adaptation theories~\cite{ben2010theory,redko2017theoretical,redko2019advances,zhang2020unsupervised} have contributed significantly towards the development of various image-based UDA approaches through solidifying the learning boundaries of UDA and providing theoretical insight over possible alignment approaches. More recent methods~\cite{zhang2019bridging,montesuma2021wasserstein} whose design aligns with the advance in domain adaptation theories have further proven their effectiveness via empirical results. However, this has not been the case for VUDA approaches, which have been designed based primarily (or even exclusively) on empirical results. While hypothesis relating to the behaviour of transferable video features have been proposed~\cite{xu2022learning,xu2022extern}, these hypothesis are also empirical-based and are not proven theoretically. Meanwhile, since the video data contains additional modalities (such as the temporal modality), existing theories developed for image-based UDA may not be extended to VUDA directly. The lack of domain adaptation theory specified for VUDA results in the poor explainability of current VUDA approaches, and further hinders the capability of VUDA approaches to be applied to data "in-the-wild". With the increasing emphasize of model explainability, breakthroughs in domain adaptation theory for VUDA would be the key for further significant improvements of VUDA methods.

\section{Conclusion}
\label{section:concl}
Video unsupervised domain adaptation (VUDA) plays a crucial role in improving video model portability and generalizability while avoiding costly data annotation by tackling the performance degradation problem under domain shift. This paper reviews the recent progress of VUDA with deep learning. We first investigate and summarize the methods for both closed-set VUDA, and non-closed-set VUDA scenarios with different constraints and assumptions of source and target domain. We observe that non-closed-set VUDA methods are more feasible in real-world applications. We further review available benchmark datasets for the various VUDA scenarios. We summarize the recent progress in VUDA research while providing insights into future VUDA research from the perspectives of leveraging multi-modal information, investigating reconstruction-based methods, and exploring other VUDA scenarios. We hope that these insights could help facilitate and promote future VUDA research, which enables robust and portable video models to be applied effectively and efficiently for real-world applications.

\bibliographystyle{ACM-Reference-Format}
\bibliography{vuda}

\end{document}